\documentclass[12pt]{article} 

\usepackage{amsfonts,amsmath,amssymb,amsthm}
\usepackage{array}
\usepackage{authblk}
\usepackage{color}
\usepackage{epsfig,fancyheadings}
\usepackage{geometry,graphicx}
\usepackage[]{hyperref} 
\usepackage{indentfirst}
\usepackage{mathtools}
\usepackage{multirow}
\usepackage{listings}
\usepackage{rotating}
\usepackage{comment}
\usepackage{booktabs}

\usepackage{sectsty,secdot}
\sectionfont{\bf\fontsize{14}{2em}\selectfont}
\subsectionfont{\bf\fontsize{12}{2em}\selectfont}

\theoremstyle{definition}

\title{{\bf\fontsize{14pt}{18pt}\selectfont AICov: An Integrative Deep Learning Framework for COVID-19 Forecasting with Population Covariates}}
\author[1]{{\fontsize{12pt}{0.5em}\selectfont Geoffrey C. Fox}}
\author[1]{{\fontsize{12pt}{18pt}\selectfont Gregor von Laszewski}}
\author[1]{{\fontsize{12pt}{0.5em}\selectfont Fugang Wang}}
\author[2,3]{{\fontsize{12pt}{0.5em}\selectfont Saumyadipta Pyne}}

\affil[1]{{\emph\fontsize{12pt}{0.5em}\selectfont Digital Science Center, Indiana University, Bloomington Indiana, USA}}
\affil[2]{{\emph\fontsize{12pt}{0.5em}\selectfont Public Health Dynamics Lab, and Department of Biostatistics, University of Pittsburgh, Pittsburgh, Pennsylvania, USA}}
\affil[3]{{\emph\fontsize{12pt}{0.5em}\selectfont Health Analytics Network, Pennsylvania, USA}}

\newcommand{\ABSTRACT}{
The COVID-19 pandemic has profound global consequences on health, economic, social, political, and almost every major aspect of human life. Therefore, it is of great importance to model COVID-19 and other pandemics in terms of the broader social contexts in which they take place. 
We present the architecture of AICov, which provides an integrative deep learning framework for COVID-19 forecasting with population covariates, some of which may serve as putative risk factors. 
We have integrated multiple different strategies into AICov, including the ability to use deep learning strategies based on LSTM and even modeling.
To demonstrate our approach, we have conducted a pilot that integrates population covariates from multiple sources. Thus, AICov not only includes data on COVID-19 cases and deaths but, more importantly, the population's socioeconomic, health and behavioral risk factors at a local level. The compiled data are fed into AICov, and thus we obtain improved prediction by integration of the data to our model as compared to one that only uses case and death data.

 }
\date{}

\begin{document}

\renewcommand{\baselinestretch}{1.07}
\renewcommand{\thefootnote}{}

\maketitle
\thispagestyle{empty}

\begin{quotation}
\centerline{\bfseries{\fontsize{14pt}{1em}\selectfont Abstract}}\par
\setlength{\parindent}{3.22cm}
{\fontsize{12pt}{1.07em}\selectfont~~\ABSTRACT}

\vspace{12pt}
\par
\end{quotation}\par

\def\thefigure{\arabic{figure}}
\def\thetable{\arabic{table}}

\renewcommand{\theequation}{\thesection.\arabic{equation}}

\fontsize{12}{1.07em}\selectfont

\setcounter{section}{0} \setcounter{equation}{0} 

\section{Introduction}

\paragraph{COVID-19.} The COVID-19 pandemic has caused unprecedented health and humanitarian crisis worldwide. It is unlike any other single phenomenon that has occurred in modern history since the end of World War II. 
The pandemic’s effects have spanned over a range that is so vast over space, and yet so condensed over time, that the dual blows of intensity and rapidity have exposed myriad systemic vulnerabilities in many societies around the world. 

The first known case was traced back to  17  November  2019. Since the emergence of a cluster of cases in Wuhan, China, on 31 December 2019, COVID-19 has spread rapidly worldwide. Just six months later, as of 23 June 2020, there are 8,993,659 COVID-19 cases and 469,587 deaths globally. On 30 January 2020, the World Health Organization (WHO) \cite{www-who} declared COVID-19 to be a Public Health Emergency of International Concern and subsequently on 11 March 2020, declared it to be pandemic in lieu of the alarming levels of spread and severity and lack of actions taken by governments around the world to curb transmission and prevent adverse outcomes.

In the United States, the first case of COVID-19 was confirmed on 20 January 2020 in Washington State, and as of 23 June 2020, there were over 2.3 million confirmed cases and 121,167 deaths \cite{www-nyt-map}. In the absence of a vaccine or treatment to effectively combat the disease, non-pharmaceutical policy interventions such as social distancing and lockdowns are recommended by health experts to prevent further transmission. To gain insights into the possible impact of such measures on COVID-19 outcomes, we depend upon the ability to accurately forecast the spread of reported cases and confirmed deaths and recoveries. Naturally, the accuracy of forecasting relies on the availability of current, reliable data and historical data to determine estimates of uncertainty.

During outbreaks of epidemics, there tends to be little to no reliable data in the beginning, and the quality of data annotation, validation, and aggregation might be uncertain. For instance, changes in clinical data entry, such as the addition of a new category {\em clinically diagnosed} to the existing {\em lab-confirmed} category, may likely have been reflected in the reporting \cite{Petropoulos2020-hh}. Initial fears about COVID-19 among the general population with regards to the trajectory of the pandemic could also affect administrative reporting. Diverse media sources and differences in local and federal policies may add to the general uncertainty about disease progression. Nevertheless, a data-driven approach to forecasting can offer valuable insights into the disease dynamics, and thereby an ability to objectively plan for the near future. 

Unlike earlier global outbreaks, during COVID-19, the current worldwide digital ecosystem allows for real-time data collection, which in turn is used by AI and deep learning systems to understand healthcare trends, model risk associations, and predict outcomes \cite{Ting2020-lw}. In addition to traditional public health surveillance strategies available to most countries, a variety of static and dynamic data types may be integrated to model the scale and dynamics of this pandemic \cite{Pyne2015-ao}. Several organizations, including the John’s Hopkins University’s Center for Systems Science and Engineering, the New York Times, the Atlantic’s COVID-19 tracking project, among others have developed real-time tracking maps for following COVID-19 cases around the world using data from the Centers for Disease Control and Prevention (CDC), WHO, and other international health agencies. Notably, the CDC receives forecasts from several modeling groups that use a wide variety of approaches ranging from SEIR (Susceptible-Exposed-Infectious-Recovered) to Agent-based to Bayesian models to understand the non-pharmaceutical policy interventions (or the lack thereof) to predict disease dynamics and its impact on human lives \cite{www-cdc-modeling-forecast}.

\paragraph{Deep Learning Framework.}
For precise and contextually relevant modeling of COVID-19 dynamics for a given population or community, we need to begin with a process of systematic generation of realistic micro (and macro) level data in the absence of their real counterparts. We will use data integration to combine community-specific health and behavioral risk factors, demographic and socioeconomic variables that are known at the local levels along with the cases and deaths data for COVID-19. Further, we use microsimulation methods to model a novel macro-level joint distribution of spatial, demographic and risk factor parameters that can provide a novel foundation to build our generative and forecasting model of COVID-19 dynamics. 

While leveraging the recently developed and increasingly popular deep learning strategies to forecast COVID-19 dynamics, we will integrate spatial, demographic, socioeconomic, health, and behavioral risk factors in our model. 

We develop a parallel computing platform called AICov standing for  {\em AI-driven Platform for COVID-19} to implement the different components of our framework and obtain the needed resources through multi-cloud interfaces in a robust manner. 

\paragraph{Data Integration.}
In this study, while we incorporate information from community and county-specific covariates to inform our forecasting model for each metropolitan area, we want to avoid the so-called ecological fallacy in drawing inferences about individual disease outcomes based on large area-level  aggregated  risk factors.   Our  main  objective  is  to  draw  attention  to  the  nuanced  roles played by pre-existing socioeconomic and other conditions of a given community that can act as  determinants  of its health outcomes and possible disparities,  especially under  the  sudden stresses to its current systems that are felt during a pandemic.

\paragraph{Paper Organization.}
The paper is structured as follows. In Section \ref{sec:lstm-theory}, we will provide a short overview of time series forecasting that we use in our framework. In Section \ref{sec:arch}, we outline our architecture in order to lower the learning curve to apply a sophisticated time series forecast while integrating risk factors in addition to the time-series data. In Section \ref{sec:data}, we describe the input data for our work.
Next, we provide our analysis of the risk factors and identify those that impact our prediction. After that, we present our conclusion.

\section{LSTM}
\label{sec:lstm-theory}

In this section, we present a short overview of the recurrent neural network algorithms used in this work (Section \ref{sec:lstm-background}).
Additionally, we provide a short introduction of the LSTM implementation as used in Keras (Section \ref{sec:keras}) that we use as a foundation of this work.

\subsection{LSTM Background}
\label{sec:lstm-background}

Within this work, we will be using a Long Short-Term Memory (LSTM) algorithm for our predictions \cite{www-keras-lstm,Hochreiter1997-dk}. An LSTM is a recurrent neural network (RNN) \cite{Rumelhart1986-li} with feedback connections allowing the use of data input sequences to predict data output sequences. LSTM's have been applied to many application areas from handwriting \cite{Graves2009-qb}, image, feature detection, and time series prediction \cite{Schmidhuber2005-oy}. LSTM's have an internal state that is used to prevent the vanishing gradient problem \cite{Hochreiter1991-mp} during the training of RNNs.

Different variants of LSTM algorithms exist. One of them we use is based on an LSTM cell that we depict in Figure \ref{fig:lstm}. 
It has an input gate, output gate, and a forget gate. The cell is maintaining how a subsequent value is calculated. This includes (a) how input values are influencing the cell via the input gate, (b) how the forget gate influences a memory value within the cell via the forget gate, and (c) how the output gate influences the output activation of the LSTM unit. A cell has a number of inputs and outputs that are weighted. During a training step, these weights are learned so that they can be reused in a prediction process between input and output sequences.

\begin{figure}[h!]
\begin{minipage}[b]{0.6\textwidth}
    \centering
    \includegraphics[width=0.6\columnwidth]{images/lstm.pdf}
    \caption{LSTM Unit Diagram}
    \label{fig:lstm}
\end{minipage}
\ \
\begin{minipage}[b]{0.3\textwidth}
\vspace{-3cm}
\begin{equation*}
\begin{split}
    f_t &= \sigma(W_f x_t + U_f h_{t-1} + b_f) \\
    i_t &= \sigma(W_i x_t + U_i h_{t-1} + b_i) \\
    o_t &= \sigma(W_o x_t + U_o h_{t-1} + b_o)\\
    \tilde{c}_t &= tanh(W_c x_t + U_c h_{t-1} + b_c)\\
    c_t &= f_t \circ c_{t-1} +i_t \circ \tilde{c}_t \\ 
    h_t &= o_t \circ tanh(c_t) 
    \label{eq:lstm_equations}
\end{split}
\end{equation*}
\caption{LSTM Equations}
\label{fig:eq}
\end{minipage}
\end{figure}

\newcommand{\Rh}{\mathbb{R}^{h}}

We have summarized the equations used in LST in Figure~\ref{fig:eq}. We denote variables in lower case and matrices in upper case letters. $W_{q}$ and $U_{q}$ contain the weights of the input and recurrent connections, where the subscript $_{q}$ denotes either the input gate $i$, the output gate $o$, the forget gate $f$ or the memory cell $c$.  It is important o note that  $c_{t}\in \mathbb {R}^{h}$ is not just one cell of one LSTM unit, but contains $h$ LSTM unit's cells, representing a vector. To summarize our variables we use
$\circ$ is the Hadaman product;
$\sigma$ is the sigmoid function;
$d$ number of input features;
$h$ number of hidden units;
$x_{t}\in \mathbb R^d$ is the input vector to the LSTM unit
$f_t\in \Rh$ is the forget gate's activation vector
$i_{t}\in \Rh$ is the input/update gate's activation vector
$o_{t}\in \Rh$ is the output gate's activation vector
$h_{t}\in \Rh$is the hidden state vector (e.g. the output vector of the LSTM unit)
$\tilde{c}_t\in \Rh$ is the cell input activation vector
$c_{t}\in \Rh$ is the cell state vector
$W\in \mathbb{R}^{h\times d}$, $U\in \mathbb{R} ^{h\times h}$  and $b\in \Rh$ are the weight matrices and bias vector parameters which need to be learned during training.

\subsection{LSTM in Keras}
\label{sec:keras}

LSTM has been available in Keras since 2015 \cite{www-keras-lstm,www-keras-api}. However, it does require some level of expertise and understanding of RNNs due to the needed manipulation of the input data that needs to be adjusted for the application. The lack of advanced reusable examples focusing on multiple input and output variables in conjunction with secondary information requires significant attention from the users. Most examples focus on very simple usage of LSTM and do not take into account the integration of other variables, such as our risk factors, in addition to time series data as input and output as we utilize in our approach.

An additional issue is that Keras depends and is distributed with Tensorflow \cite{www-tensorflow}. Although Tensorflow claims to be the most popular Deep Learning framework, it is surprising that its adoption within modern operating systems has recently lacked behind other frameworks such as PyTorch \cite{www-pytorch}. Furthermore, it is also dependent on NVIDIAs drivers and ports to the OS that presently lacks behind. 
Although a container image is presented, the examples demonstrated for using it are limited.

\section{Architecture}
\label{sec:arch}

We discuss and present our requirements that lead to our architecture framework and how we address them.

The motivation for an architecture is based on our experience with deep learning toolkits such as Keras and PyTorch. Although these systems provide the necessary APIs to integrate time series solutions for data sets, they target primarily more general user communities. In particular, they do not consider existing data sets or specific needs and analysis options to evaluate them for the use within COVID-19 risk factor integration. The availability of such a tool would make it more suitable for scientists to reuse such a framework while focusing on the analysis part. We list in more detail some of the requirements in a more formal way that motivate our architectural design:

\begin{description}

\item[Easy to Use.] One of our main criteria is that the framework must be easy to use, and extensions are provided that address usability on the API as well as the user interface level. {\bf Solution:} To address these issues, we are devising a specialized time series API for COVID-19 that integrates common tasks such as automated filtering and normalization of the data.

\item[Interactive.] Due to the experimental nature of analyzing the data, the framework must support the exploration of the data in an interactive fashion. {\bf Solution:} As many data scientists use Jupyter Notebooks to integrate their work interactively, our framework must also integrate easily into such notebooks. Jupyter notebooks allow easy modification of the analysis workflow through the ease of Python as programming language. At the same time, they provide a level of sophistication to formulate easily scientific analytics workflows with the help of Python.

\item[Expandable API.] As we assume that new models and other analysis algorithms are developed over time, our framework must allow the integration of these APIs. {\bf Solution:} For this reason, we develop an abstraction {\bf API} for data, analysis, and metadata parameter adjustments. Furthermore, external services can be integrated with the help of {\bf REST} services.

\item[Data Source Integration.] The integration of various data sources is a key part of our framework. {\bf Solution:} To achieve this goal, we provide a number of abstractions and data manipulation functions to extract needed data from established sources. Furthermore, the data can be combined, and additional data wrangling activities through outside groups can be integrated.
Through such abstraction, it will be possible to replace, add, and correct data sets that are part of the analysis. This allows us to compare different results created from different data sets. Moreover, we include integrate data integration capabilities directly in our abstractions to easily and quickly resue the enhanced data. An important example is that the data is not static throughout the disease but is continuously updated. This also has a significant impact on our analytics algorithms.

\item[Automated Recalculation of the Analytics.] As the data can change, a previous analysis may have to be rerun to stay up to date. {\bf Solution:} Our forecast is not only a single script, but includes a mechanism to register multiple analysis workflows that are automatically rerun once new data is made available to the system. Old results are maintained. Metadata with these runs can retrieve the versioned data sets used for the calculation and store the versioned analytics workflow.

\item[Flexible Model definition.] We want to experiment with a variety of models that need to be easily and flexibly be definable as workflows. {\bf Solution:} We need functional abstraction definitions that allow us to define the models ourselves or integrate third party models describing the spread of the disease. 

\item[Deep Learning Forecast.] As the forecasting models depend on changing data that are made available daily, it is important that we rapidly adopt to the new data availability. Many models can be applied to this, including moving averages based models, or models that are purely derived from deep learning. Our framework will allow the integration of both. Furthermore, in our deep learning framework, we will integrate an automated search for important risk factors.

\item[Model Orchestration.] To coordinate the different model predictions and the generation of the deep learning forecasts, we need to be able to orchestrate them while applying a number of parameters. {\bf Solution:} the framework includes the ability to integrate parameter sweeps to, for example, identify hyperparameters, or integrate different data sets as parameters to identify suitable forecast models that deliver the best fit.

\item[Compute Resource Mapping.] As others want to use the framework, we need to integrate a flexible resource utilization framework. {\bf Solution:} We will leverage from our earlier work and utilize through Cloudmesh and NIST NBDIF definitions cloud services into the architecture. This will include containers but also infrastructure in the cloud and locally that provides GPUs that can be leveraged within our architecture.

\item[Interface.] As we have a wide variety of users, we need to enable interfaces used by the various communities. {\bf Solution:} As discussed, we will provide a number of API's, REST services and make them assailable via Jupyter Notebooks. In addition, we will develop some custom widgets for the notebooks that are specifically targeted towards the data integration, parameter manipulation, and visualization of the results 
\end{description}

\begin{figure}[h!]
    \centering
    \includegraphics[width=0.6\columnwidth]{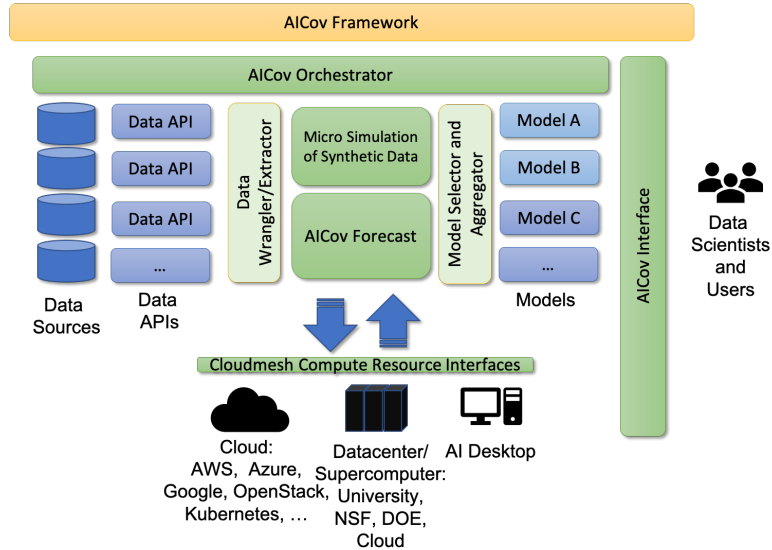}
    \caption{AICov Architecture}
    \label{fig:arch}
\end{figure}

Putting those requirements together results in an architecture as shown in Figure \ref{fig:arch}.

 The architecture allows the inclusion of new data sources and models. An orchestrator enables the modification of parameters and the selection of data sources and models used in the analysis. The framework can be accessed via the interface layer that includes APIs, REST, and Jupyter Notebooks. Compute interfaces are provided either by direct access of GPUs from local resources or through the staging of automatically generated REST services via our Cloudmesh REST Service integrator. Additionally, we can also access through abstractions cloud services that are offered by the various cloud providers and are targeting our application domain. Different models can be integrated into our framework to verify our forecast or enhance our forecast based on models and data selected by the users.

\section{Data for the COVID-19 Analysis}
\label{sec:data}

While the U.S. has seen an overall case fatality rate of 5.14\%, some U.S. locations have experienced a disproportionate number of deaths compared to others. This disparity of deaths among different communities could be attributed to a diversity of local risk factors and socioeconomic determinants of health that now constitute a very active area of investigation. The researchers are studying the effects of different community-specific socioeconomic variables, underlying health conditions and comorbidities, infrastructure, and systemic responses.
Given the complexity of such underlying conditions that can determine individual disease outcomes,  where and under what conditions people get infected by the virus, and then recover or die take place not in isolation but rather in an interplay of diverse factors that are characteristic to a local community, which need to be synthesized. In other words, a basic requirement for effective modeling of COVID-19 dynamics is that the information regarding a particular community’s vulnerabilities and strengths pertaining to the disease must be as comprehensive and integrative as possible. 
The regular reporting of accurate Coronavirus outbreaks data from local health departments has been understandably difficult, especially in the areas that are hardest hit. We use data on community-specific health and behavioral risk factors aggregated at county level from the large-scale longitudinal CDC BRFSS surveys \cite{www-cdc-brfss}, and demographic and socioeconomic data from the US Census Bureau that are known at the local levels along with the cases and deaths data compiled by NCHS \cite{www-cdc-nchs} and integrate them with other data sources.

\subsection{Data collection and processing}

Among the different sources of data on daily cases and deaths data collected and aggregated at different spatial levels, especially for the US, many are either freely available repositories such as GitHub or accessible via specifically developed online resources such as the New York Times~\cite{github-nytimes} and the Johns Hopkins Coronavirus Resource Center~\cite{www-jh-covid19}. 
For this study, we generated two datasets based on the latter resource upon our checking for the locations and times at which cumulative counts of COVID-19 cases and deaths on any given day were no less than those on the previous day. This led us to compile daily coronavirus cases and deaths data for 345 US counties from 32 states and the District of Columbia and matched by five-digit FIPS code or county name to dynamic and static variables from additional data sources.

The static data in this repository was collected from the US Census Bureau, and the Centers for Disease Control and Prevention (CDC). The different fields cover the domains of behavioral and health risk factors, hospital capacity, and socioeconomic and demographic conditions. Data on health outcomes, prevention measures, and unhealthy behaviors were extracted from the CDC's 500 Cities Local Data for Better Health program \cite{www-cdc-chronic-data} in the form of crude or age-adjusted prevalence of conditions such as respiratory disease, obesity or smoking. These values were aggregated from census tract to the county level using the first five digits of the eleven-digit tract FIPS code. Because a single county may consist of multiple individual cities, we include the list of all city labels within each aggregate group to represent a greater metropolitan area. 110 of such metropolitan areas that had more than 500 reported cases of COVID-19 by April 15, 2020, were selected for this study.

Notably, a greater metropolitan area such as New York City could be spread across more than one counties, which can lead to the complexity of aggregating counts and other variables. Time series data for the 5 individual counties (boroughs) in New York City were not available in the Johns Hopkins CSSE dataset. Rather, the total number of cases and deaths for the entire city of New York are reported and assigned the FIPS code of New York County (Manhattan). To ensure consistent geography across the static and dynamic data, we compute the population-weighted sum or median of each covariate over the assigned counties.

Demographic variables were gathered from the Census QuickFacts~\cite{www-census} online resource using an automated web scraping algorithm and cover relevant areas such as age, race, income and population density. Additional socioeconomic variables include the Gini Index, which measures economic inequality and CDC Social Vulnerability Index (SVI)~\cite{www-cdc-svi}. The SVI was created to guide public health officials and disaster response efforts by identifying the communities across the United States most likely to need support during a crisis. Census tracts and corresponding counties are ranked across 15 social factors, which are grouped into four themes: socioeconomic status, household composition and disability, minority status and language, and housing and transportation.

Lastly, the total number of general acute care, critical access, and military hospitals within each county are included in the data. Such variables included the number of relevant hospitals per county, and the estimated number of beds (total known bed counts added to the number of hospitals in the county with missing data times the average number of beds per hospital in that state) per 1,000 people using the American Hospital Directory~\cite{www-ahd}.
 
\section{Analysis}

\subsection{Covariance Analysis}

\begin{figure}[p]
    \centering
    \includegraphics[width=0.8\textwidth]{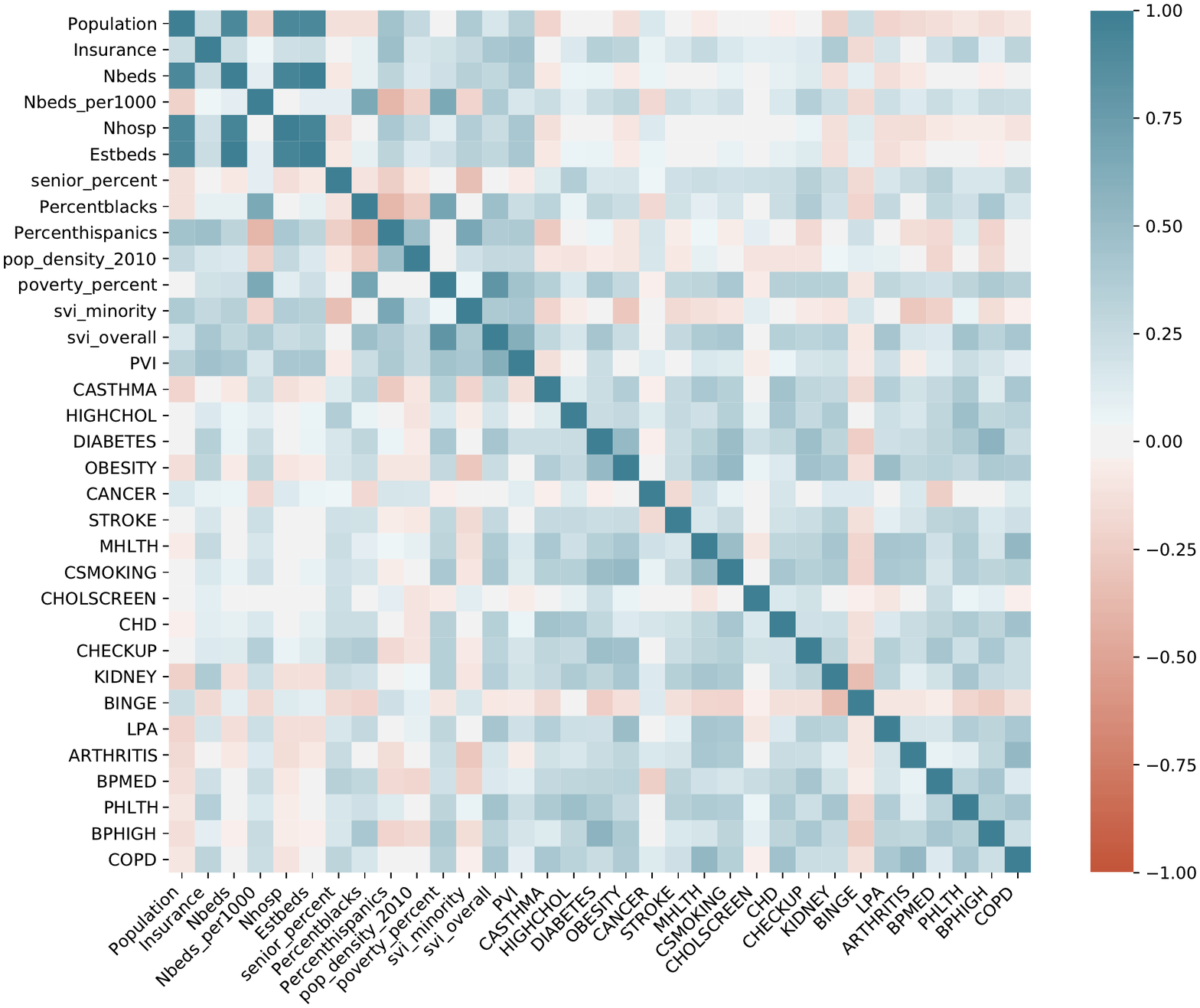}
    \caption{Covariance Analysis Heatmap of Socio-economic and Health Risk Factors}
    \label{fig:cov-heatmap-1}
    
    \bigskip
    \centering
    \includegraphics[width=0.8\textwidth]{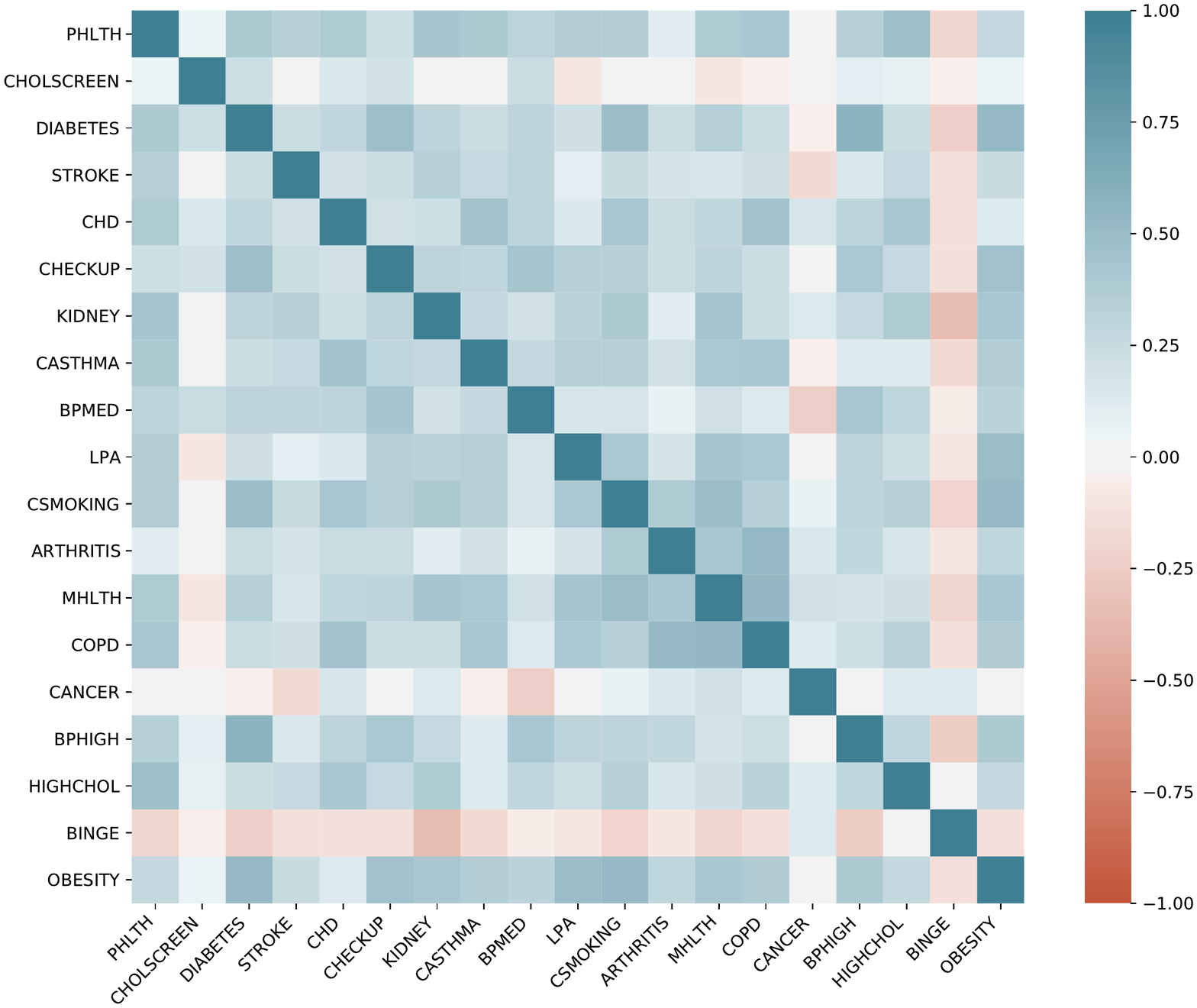}
    \caption{Covariance Analysis Heatmap of Health Related Risk Factors}
    \label{fig:cov-heatmap-2}
\end{figure}

\begin{figure}[p]
    \centering
    \includegraphics[width=1.0\textwidth]{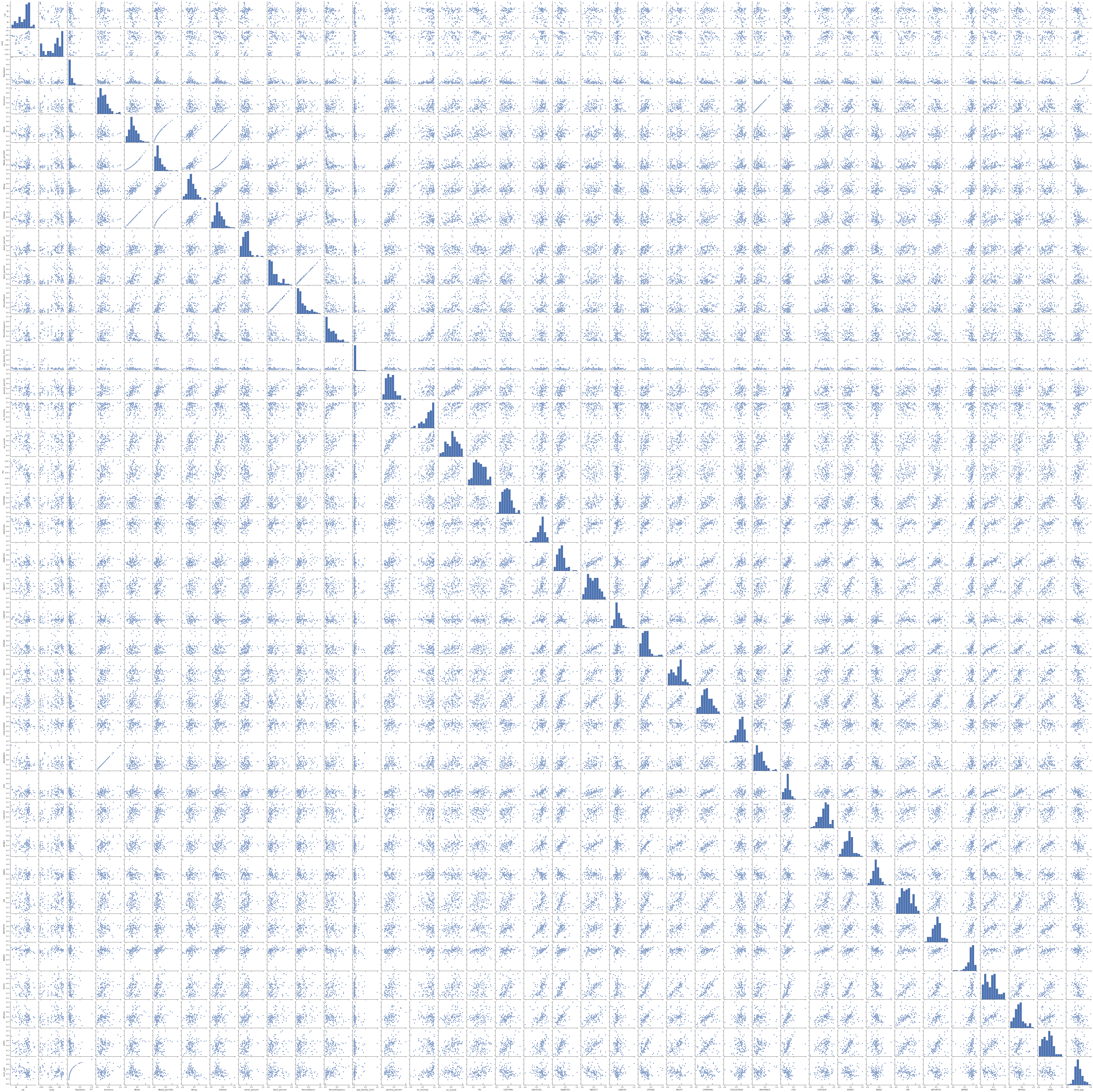}
    \vspace{-1cm}
    \caption{Covariance Analysis Pairplot of Socio-economic and Health Risk Factors.
        An online version of the image is available at \url{https://github.com/laszewski/papers/blob/master/vonLaszewski-covid19-cov-factors.pdf} 
    }
    \label{fig:cov-pairplot}
\end{figure}

For each FIPS we have a number of risk factors for which we have values. These values can be correlated with each other through a correlation analysis.   To provide an initial idea how the risk factors are correlated with each other we showcase in Figure \ref{fig:cov-heatmap-1}  and \ref{fig:cov-heatmap-2} in the scale of -1.0 to 1.0. While the first figure shows all variables, we display in the second figure, only the variables related to health measurements and not social-economic factors.
In addition, we show in Figure \ref{fig:cov-pairplot} the more detailed pair plots. However, an online version of this plot is available, as indicated in the caption, as the information for the later is very condensed. 

We observe that most social-economic variables are more strongly correlated with each other than the health variables. This is evident from the plot due to its coloring that is darker than the rest of Figure \ref{fig:cov-heatmap-1}. Obviously, the number of available hospitals vs. the number of estimated beds and the population are strongly correlated.  

When looking at the health-related risk factors, we, for example, see that diabetes and high blood pressure are correlated with each other. This reflects observations from health care providers that have correlated these two variables. A similar correlation we see in the correlation between obesity and diabetes. Having such detailed information at hand makes it now suitable to integrate such health-related factors to COVID-19 cases and death. 
This is conducted through two different pilot studies.

In the first, we relate the risk factors based on information we predict for the maximum of the cases and death from the model we used for the given time period. The second is when we use our model predictions and find the best predictions while minimizing the cumulative error for the given prediction time period. To be complete, we have included all risk factors in our analysis rather than to restrict them based on our correlation matrix.

\subsection{Empirical Fits} \label{sec:emperical}

\begin{figure}[!h]
    \centering
    \includegraphics{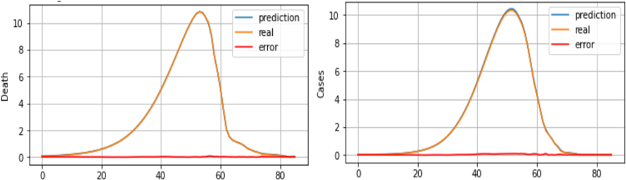}
    \caption{Empirical fits can be very accurately predicted with LSTM}
    \label{fig:magic-1}
\end{figure}

One of our goals is to integrate datasets describing risk factors in particular but social and environmental determinants of health in general and determine their combinations that are relevant for our model predictions. This will allow us to model the disease dynamics in a much broader and powerful framework. We not only focus on time-dependent and spatial instances of the disease, but also on the underlying population characteristics. To establish the feasibility of our approach, we have, in our analysis, compared them to epidemiological models with empirical fits \cite{marsland20-covid-paper}. The time series were 100 days long, and a multi-layered Long Short-Term Memory (LSTM) [30] recurrent network was used. It differed by learning not only from the demographics (fixed data for each city) and time-dependent data but by integrating the population model for the underlying prediction, as shown in Fig. 2. Such model predictive integration capability is important in any application with multiple time scales. For example, it will allow us to integrate multiscale time effects into the forecast, which could address the combination of general forecasts and the next time steps of choice such as days, weeks, months, or longer. For this pilot, we used 37 of the 110 cities with reliable empirical (not deep learning) fits for the case and death data up to April 15, 2020 \cite{marsland20-covid-paper}. Our pilot identified a sophisticated single deep learning time evolution operator that can describe these 37  separate datasets, and smooth fitted data leads to very accurate deep learning descriptions as depicted in both Figure \ref{fig:magic-1}. These models can be reused to conduct further studies to identify correlations towards particular risk factors, as shown in Section \ref{sec:max-analysis}.

\newcommand{\mysize}{0.9}

\begin{figure}[p]
    \centering
    \begin{minipage}{0.45\textwidth}   

    \includegraphics[width=\mysize\textwidth]{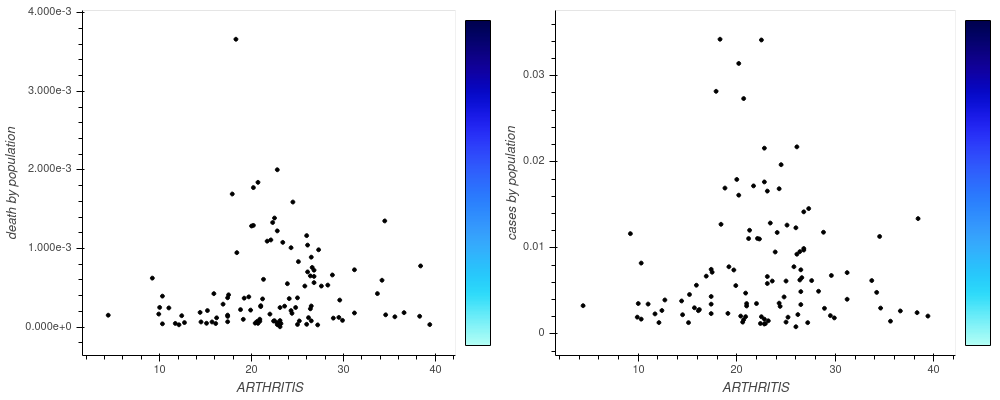}
    \includegraphics[width=\mysize\textwidth]{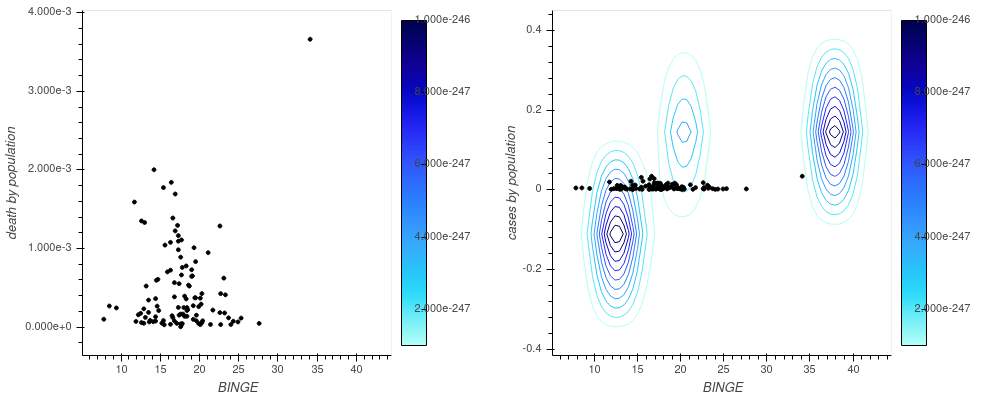}
    \includegraphics[width=\mysize\textwidth]{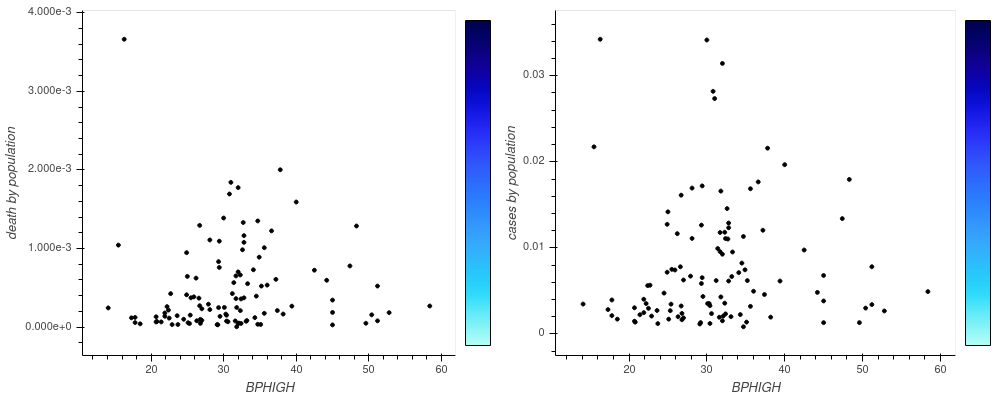}
    \includegraphics[width=\mysize\textwidth]{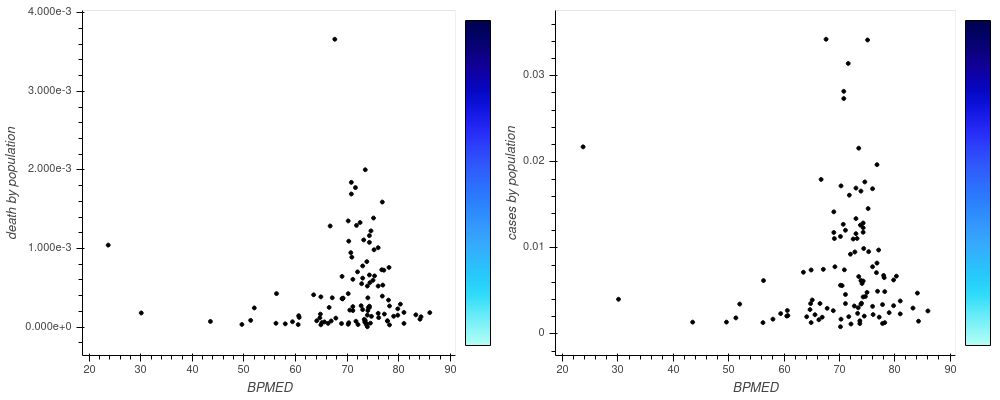}
    \includegraphics[width=\mysize\textwidth]{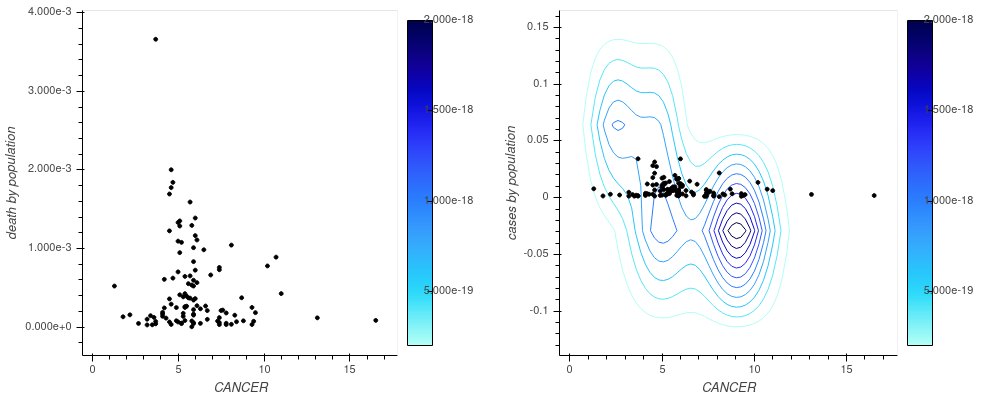}
    \includegraphics[width=\mysize\textwidth]{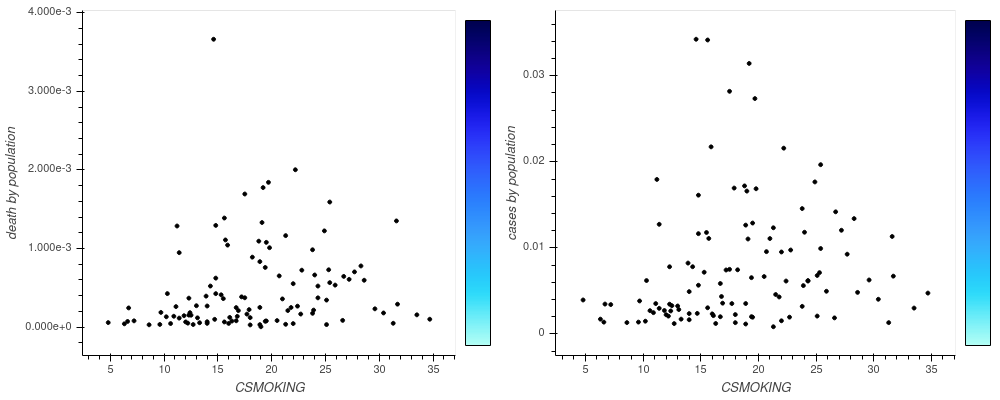}
    \includegraphics[width=\mysize\textwidth]{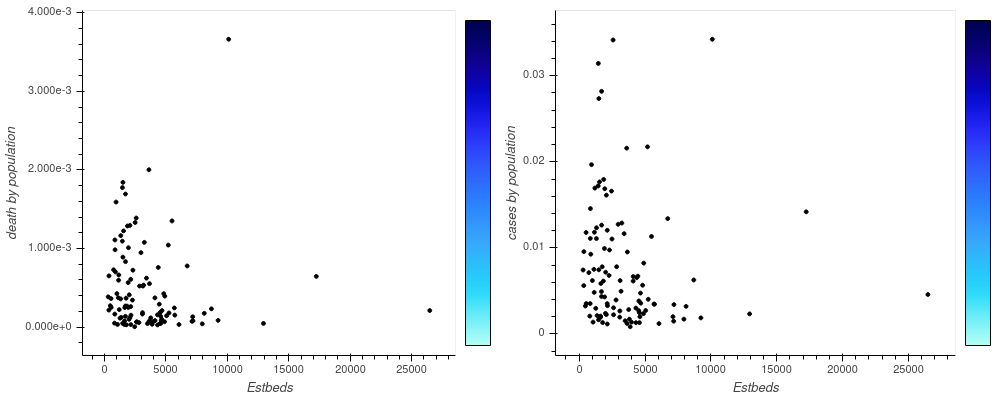}
    \includegraphics[width=\mysize\textwidth]{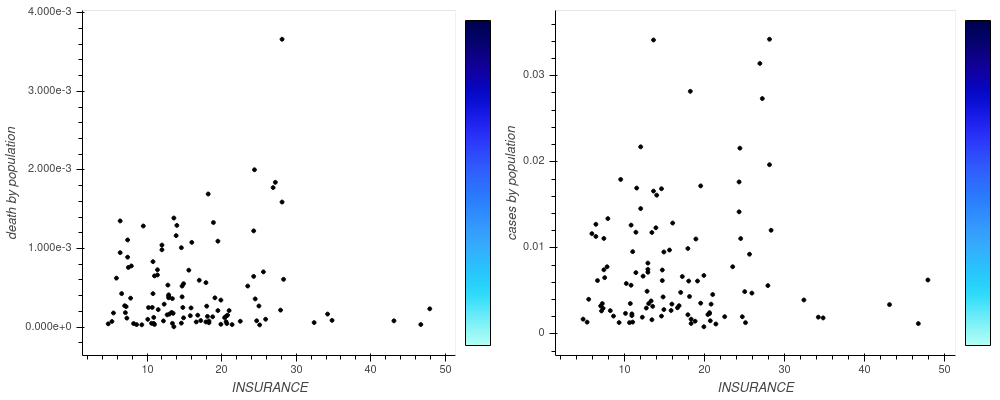}

    \end{minipage}
    \ \hfill\vline\hfill \
    \begin{minipage}{0.45\textwidth} 

    \includegraphics[width=\mysize\textwidth]{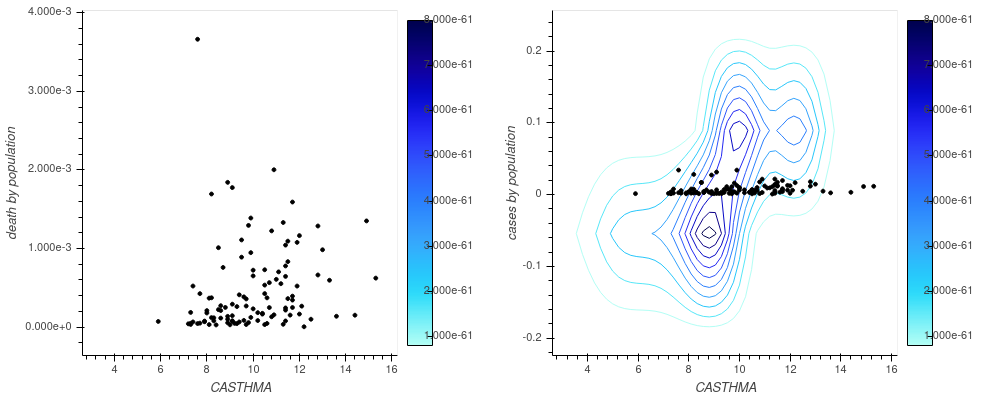}
    \includegraphics[width=\mysize\textwidth]{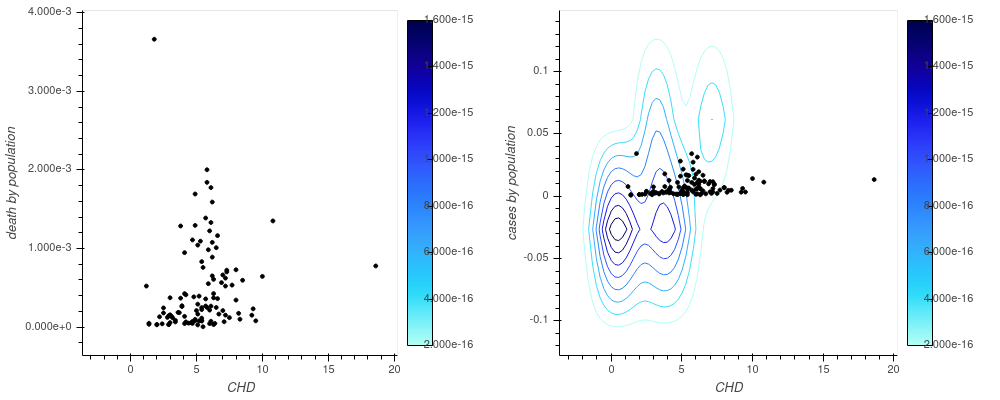}
    \includegraphics[width=\mysize\textwidth]{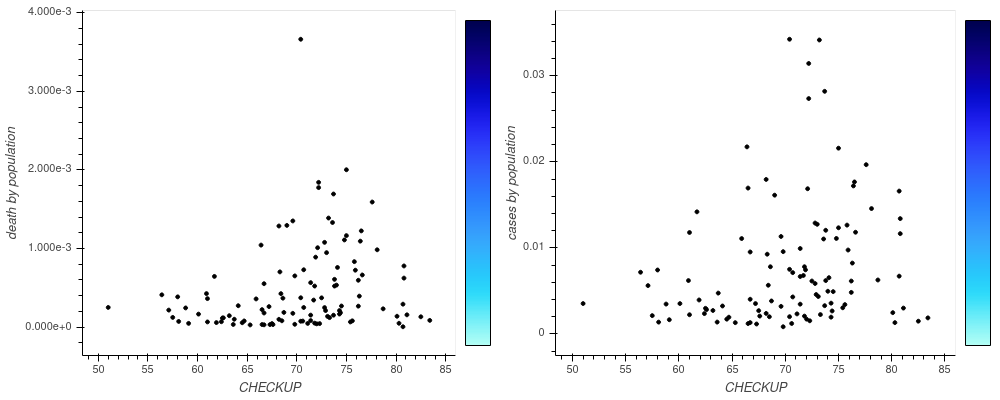}
    \includegraphics[width=\mysize\textwidth]{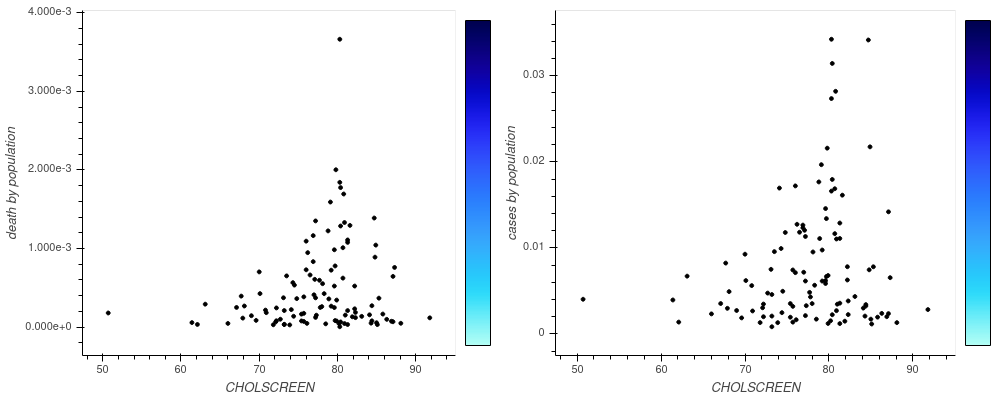}
    \includegraphics[width=\mysize\textwidth]{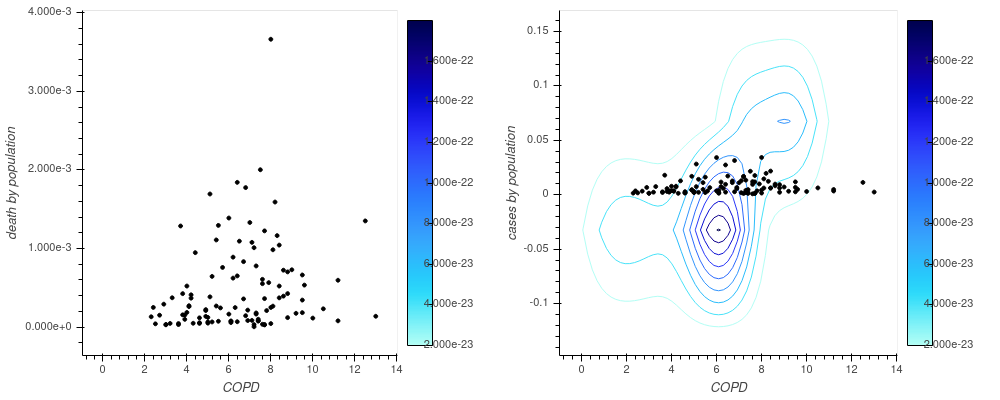}
    \includegraphics[width=\mysize\textwidth]{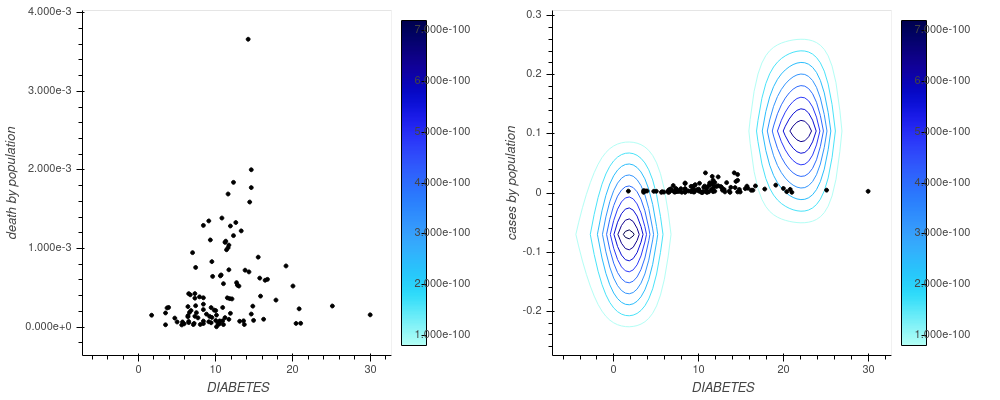}
    \includegraphics[width=\mysize\textwidth]{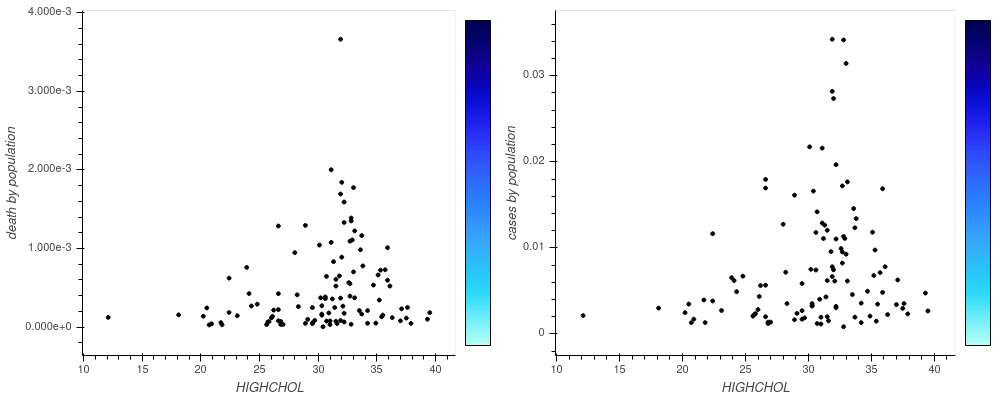}
    \includegraphics[width=\mysize\textwidth]{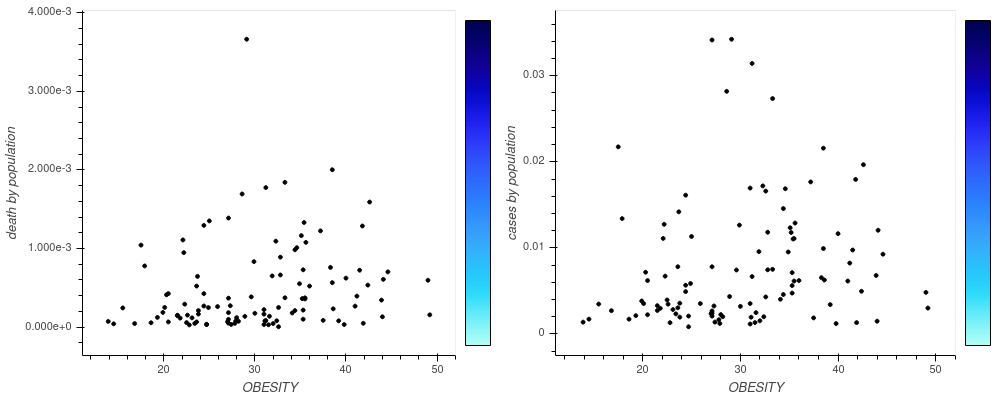}

    \end{minipage}
    \caption{Bivariate plot of risk factors when looking at maximum of cases and death in the empirical fit}
    \label{fig:risk-1}
\end{figure}

\begin{figure}
    \centering

    \begin{minipage}{0.45\textwidth}

    \includegraphics[width=\mysize\textwidth]{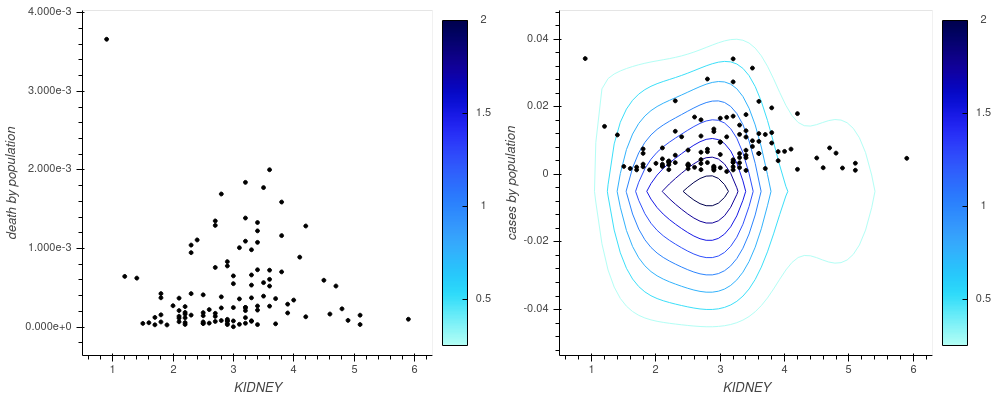}
    \includegraphics[width=\mysize\textwidth]{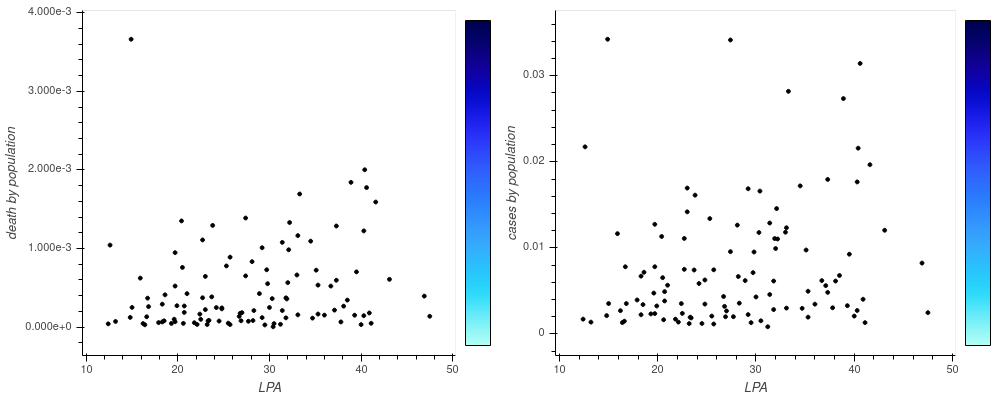}
    \includegraphics[width=\mysize\textwidth]{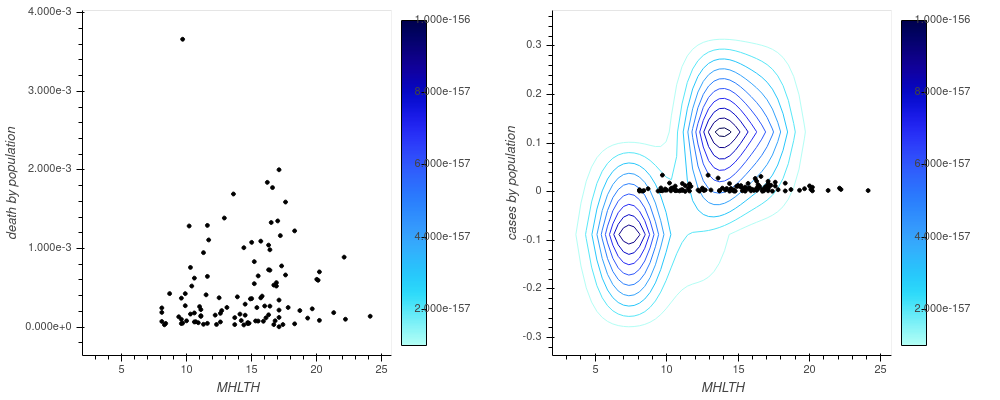}
    \includegraphics[width=\mysize\textwidth]{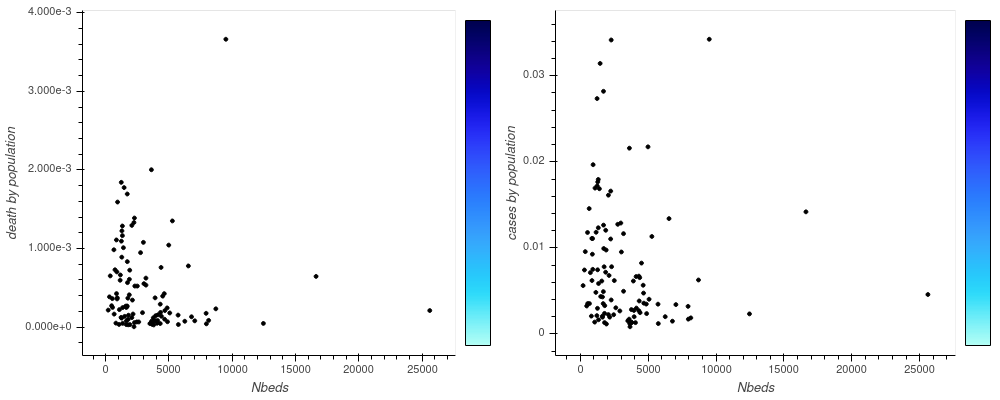}
    \includegraphics[width=\mysize\textwidth]{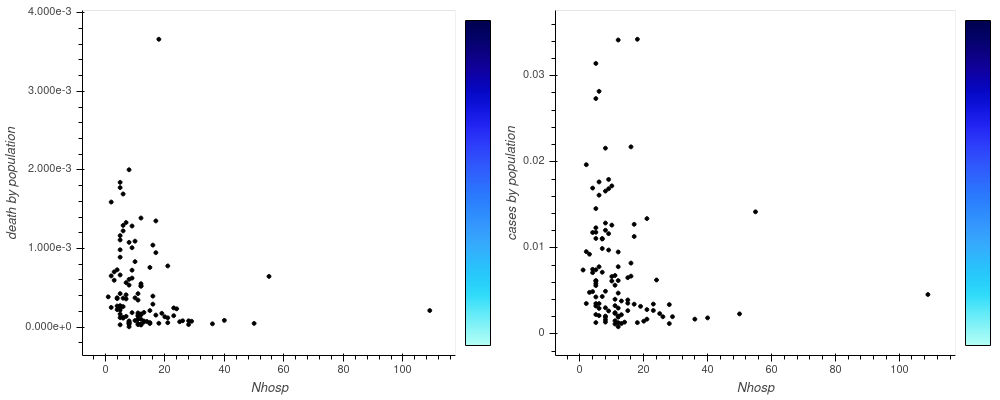}
    \includegraphics[width=\mysize\textwidth]{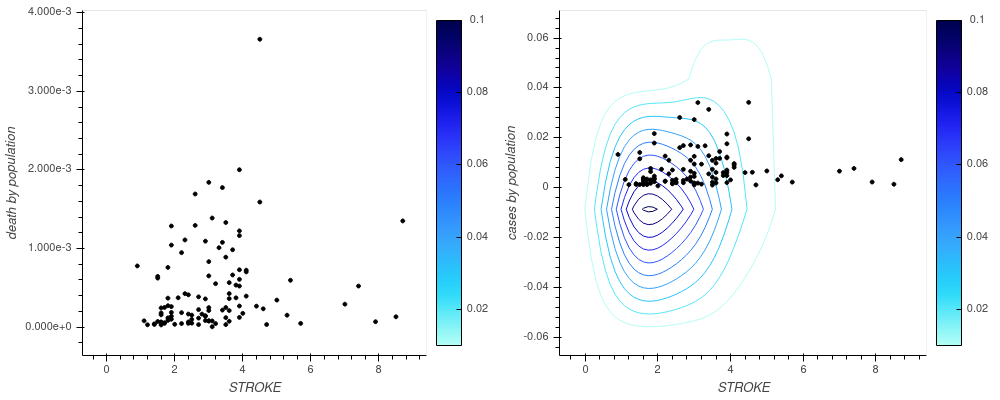}
    \includegraphics[width=\mysize\textwidth]{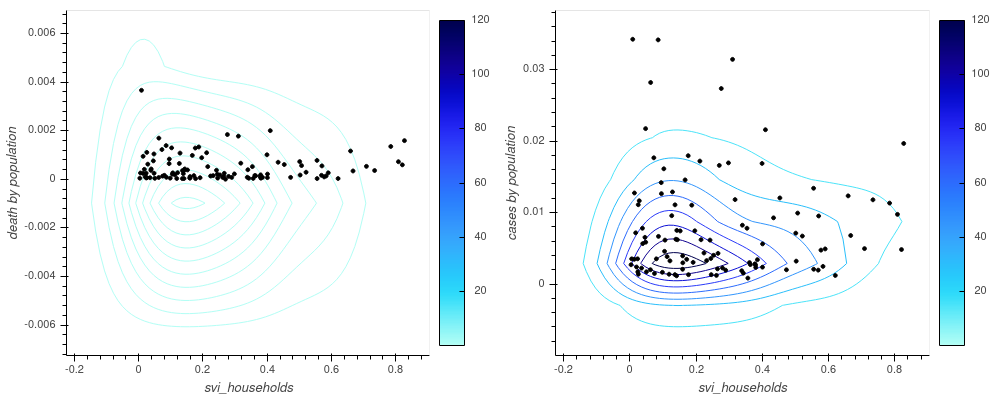}

    \end{minipage}
    \ \hfill\vline\hfill \
    \begin{minipage}{0.45\textwidth} 

    \includegraphics[width=\mysize\textwidth]{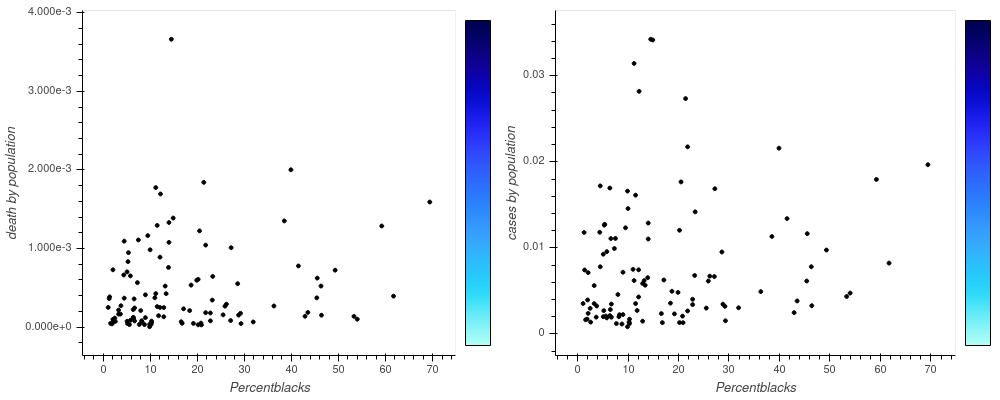}
    \includegraphics[width=\mysize\textwidth]{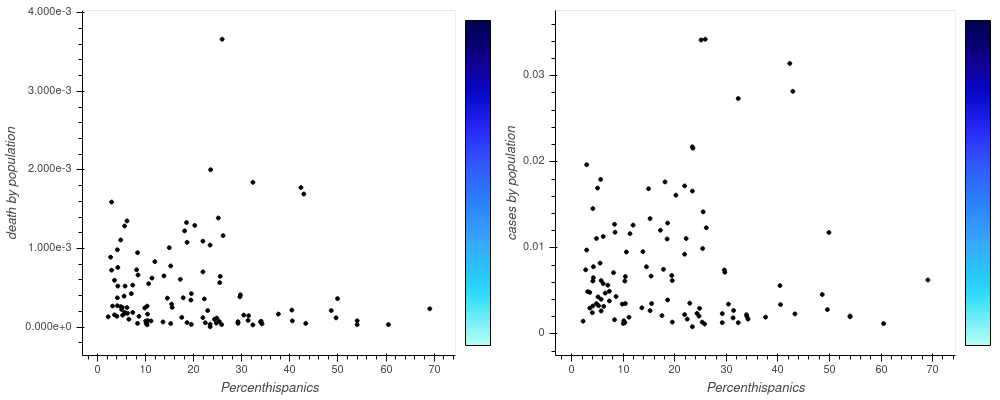}
    \includegraphics[width=\mysize\textwidth]{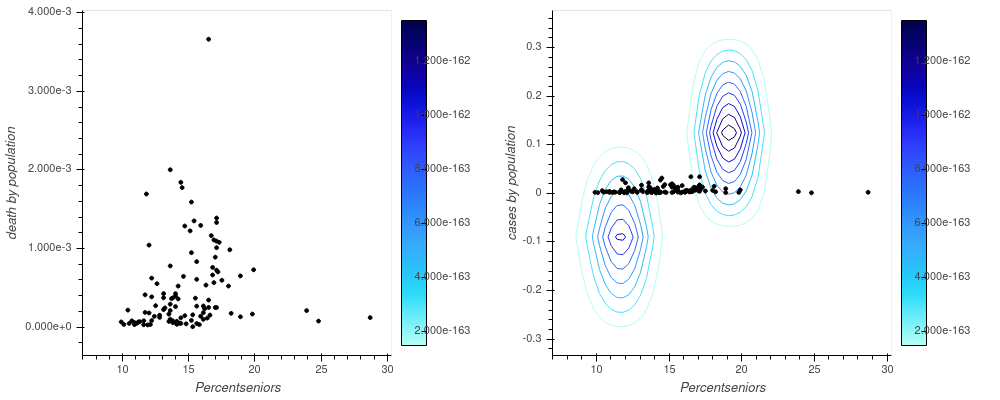}
    \includegraphics[width=\mysize\textwidth]{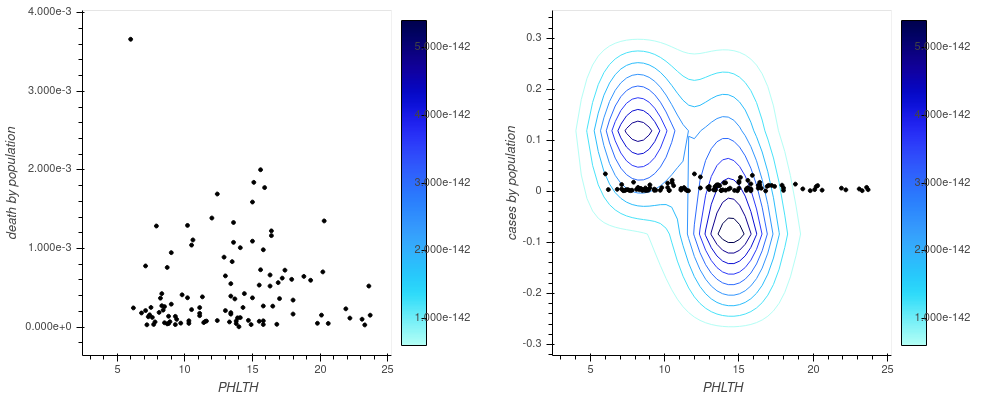}
    \includegraphics[width=\mysize\textwidth]{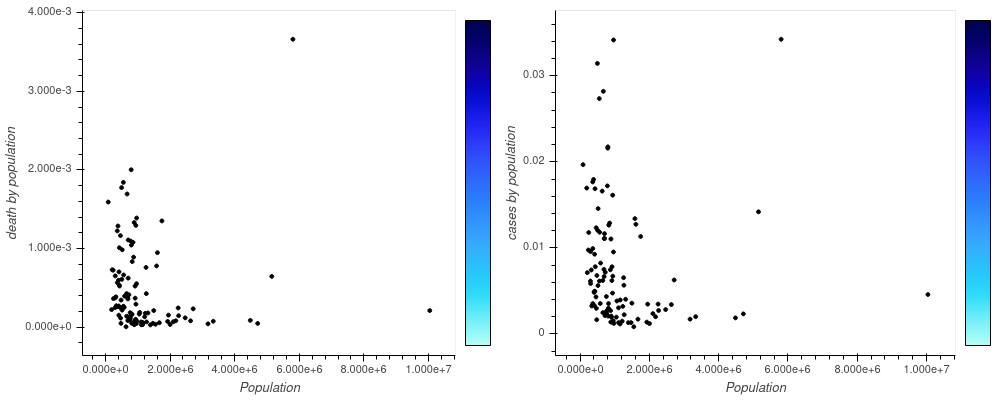}
    \includegraphics[width=\mysize\textwidth]{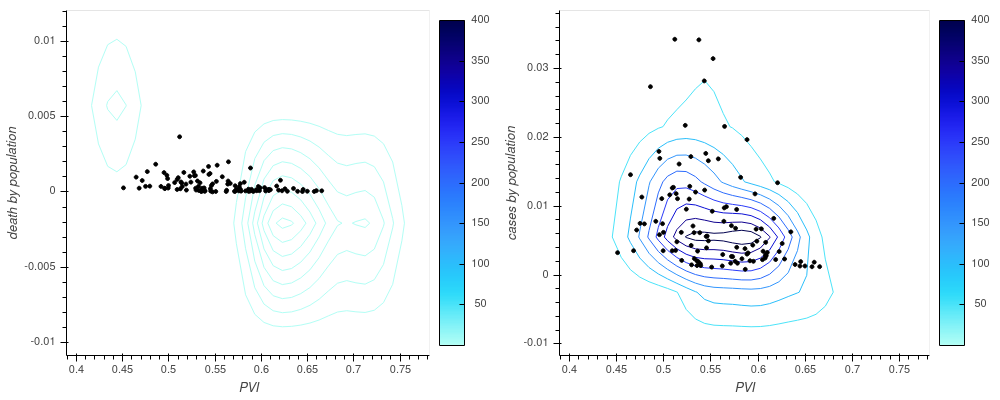}
    \includegraphics[width=\mysize\textwidth]{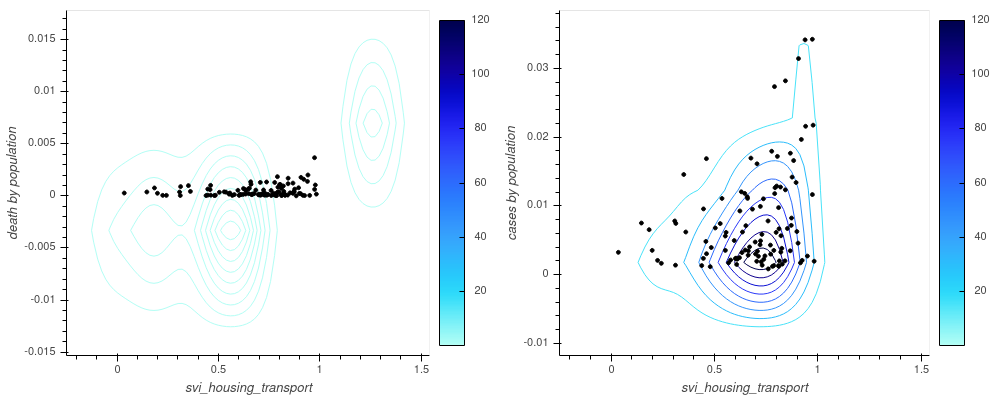}
    \includegraphics[width=\mysize\textwidth]{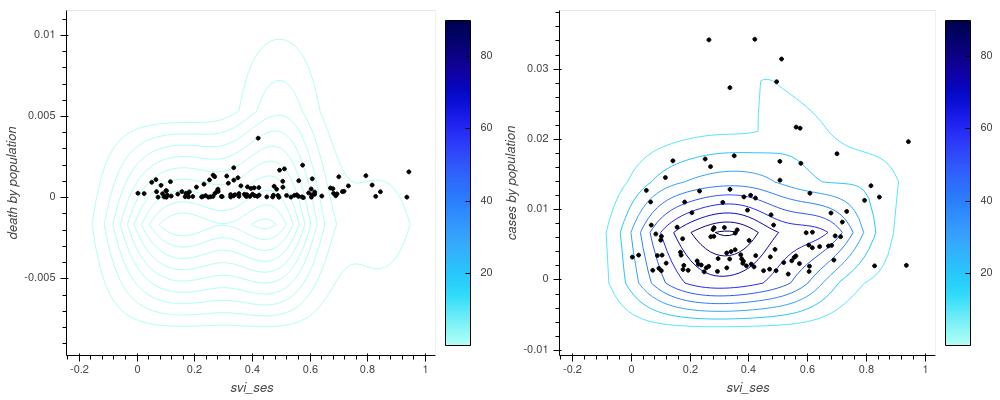}

    \end{minipage}
    \caption{Bivariate plot of risk factors when looking at maximum of cases and death  in the empirical fit}
    \label{fig:risk-2}
\end{figure}

\subsection{Correlation to the Maximum}
\label{sec:max-analysis}

One of the questions that arise is what risk factors, in addition to the cases and death, influence the prediction. In this study, we are not focusing on the cause, but like to identify suitable candidates to analyze further. For this reason, we look at the risk factors that are listed in Table \ref{tab:risk-factors}. To identify a first set of suitable candidates, we used the model as described in Section \ref{sec:emperical} and identified for each risk factor the maximum for death and for cases for each of the 110 FIPS. We display the bivariate plots for each of the risk factors in Figures~\ref{fig:risk-1} and \ref{fig:risk-2}. We can see two types of effects in these bivariate plots.

\begin{description}

\item[Increased numbers with increased population when a risk factor is present.]~\\
When looking at the variables we see for example that the risk factor STROKE has a stronger correlation to the  death per population, then it has for the number of cases. Although we are not commenting on the causality, it is intuitive for this variable to think about that a person with prior history of a STROKE has an increased risk to dying in case of an infection. However, we also see an increased number for the cases. THis indicates that STROKE is a potential important RISK factor and should be considered when conducting predictions based on death and cases.

\item[Value dependency categories for risk factors.] ~\\
When looking at the bivariate plots for death and for cases for a particular risk factor, we can identify trends within the plot that cluster some of the 110 FIPS into different categories. In the most simple aspect of this we can find liner trends that define the upper and lower bound of the risk factor bivariate plot. For example, when we look at the encompassing trend lines for cases vs. death by population, we find on the one hand FIPS for 17031,6037, 36119, 12086, representing  Cook, Illinois, US,     Los Angeles, California, US and  Miami-Dade, Florida with the PVI indices 0.582,    0.63,    0.635. On the other hand, we find Wayne, Michigan,     Essex, New Jersey,      Hartford, Connecticut,     Oakland, Michigan,     Macomb, Michigan,      Norfolk, Massachusetts. Which have  a PVI index of 0.477,     0.564,     0.527,     0.471,     0.499,     0.498 respectively. Hence, we can assume that a factor falling into different categories such as high PVI vs. low PVI have an impact on the cases and death. Associating them with the maximum cases and death helps to identify such categories. In our follow up analysis, we will identify if this is indeed the case.

\end{description}

\section{LSTM Prediction}
\label{sec:lstm-prediction}

As we wont to evaluate the influence of the risk factors on our prediction framework using not only models but deep learning algorithms, we have devised a parameter sweep framework based on LSTM that conducts the prediction on the 110 cities while also considering the risk factors as listed in Table \ref{tab:risk-factors}.
We ask:

\begin{quote}
{\em Can we increase accuracy of prediction while integration risk factors?}
\end{quote}

For this experiment, we have chosen a start date range between 2020-02-01 and 
2020-05-25.  For the LSTM we have selected the activation function based on 
Rectified Linear Units (RELU). A drop out rate of 0.2 is used, we use 16 input nodes, and have a batch size of 110. The maximum epoch is chosen to be 200. 
the number of nodes internally is 32. The total number of Samples is maximal 12540 when using 3 days as input length. We use a two-layer LSTM, as shown in Table \ref{tab:model}. Within our system we have the ability to set a number of input days. For this experiment, we have chosen three different input length, namely 3, 4, and 5 days.

\begin{table}[p]
    \caption{The deep learning parameters used for the model sweep creation in as exported by Keras using 1 risk factor. The activation function is RELU and the recurrent activation function is sigmoid.}
    \label{tab:model}
    \bigskip

    \centering
    \resizebox{1.0\textwidth}{!}{ 
    \begin{footnotesize}
    \begin{tabular}{llp{2cm}p{2cm}p{2cm}p{2cm}p{2cm}p{2cm}p{2cm}p{2cm}p{2cm}}
    \toprule 
     Layer & (type)   &
     Output Shape &  Parameters for 0 risk factor &  
     Output Shape &  Parameters for 1 risk factor &
     Output Shape &  Parameters for 33 risk factors &
     Output Shape &  Parameters for 37 risk factors\\
    \midrule
    dense & (Dense)    & (None,1,32) & 96   &  (None,1,32)  & 128  &  (None,5,32) & 1152 & (None,5,32) & 1280 \\
    lstm & (LSTM)      & (None,1,16) & 3136 &  (None,1,16)  & 3136 &  (None,5,16) & 3136 & (None,5,16) & 3136 \\
    lstm\_1 & (LSTM)   & (None,16)  & 2112  &  (None,16)    & 2112 &  (None,16)   & 2112 & (None,16)   & 2112 \\
    dense\_1 & (Dense) & (None,32)  &  544  &  (None,32)    & 544  &  (None,32)   &  544 & (None,32)   &  544 \\
    dense\_2 & (Dense) & (None,30)  &  990  &  (None,30)    & 990  &  (None,30)   &  990 & (None,30)   &  990 \\
    \bottomrule
    \end{tabular}
    \end{footnotesize}
    }
    ~\\
    \bigskip
    Sequential model with a total of 6,910 trainable parameters in case we include one risk factor. In the case, we only include no risk factor, it is 6878, and when we include 33 risk factors, it is 7934.  
\end{table}

\begin{figure}[!p]
    \centering
    \includegraphics[width=1.0\textwidth]{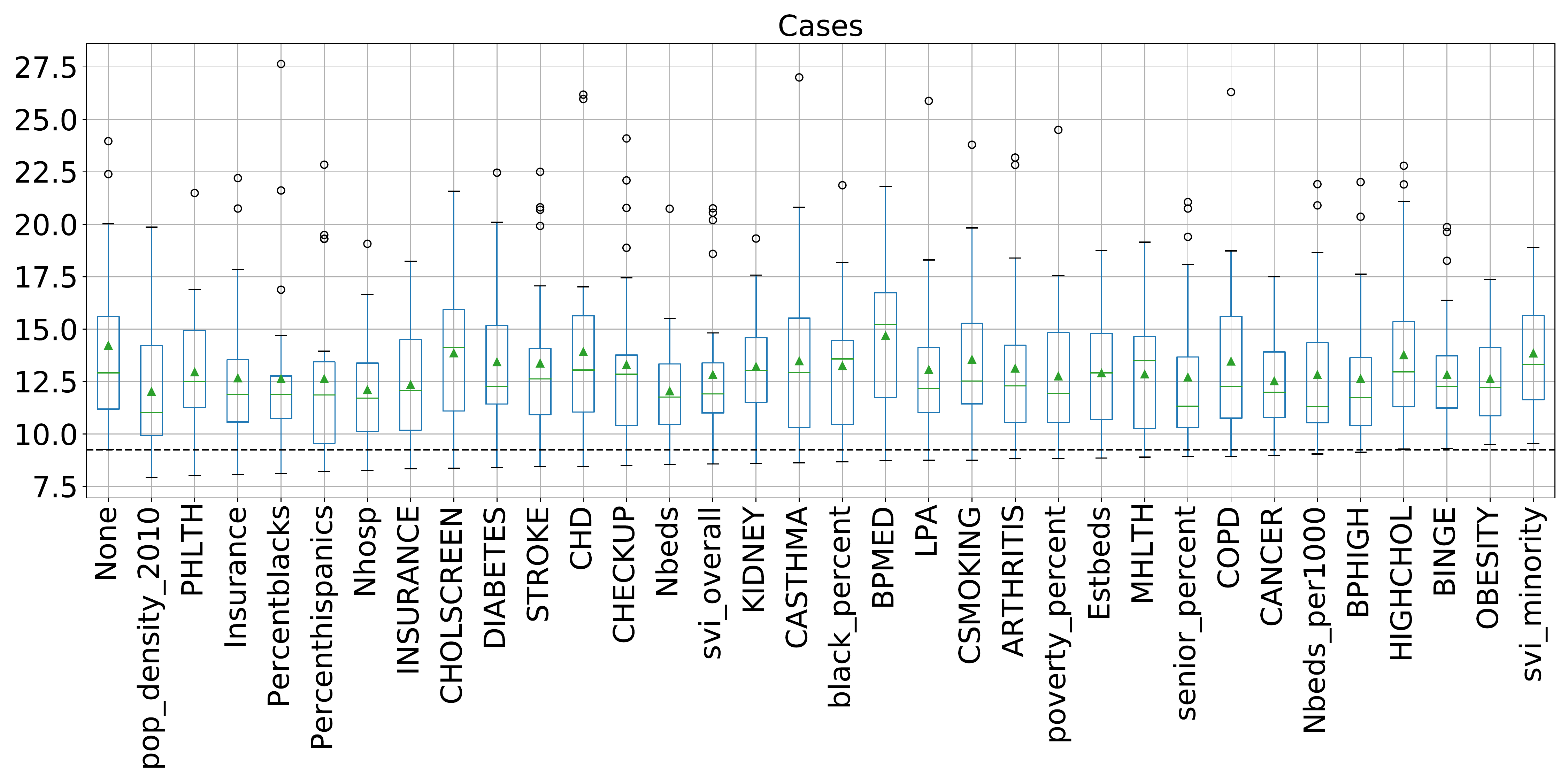}
    \vspace{-1cm}
    \caption{Influence of risk factors on the accuracy of the prediction of COVID-19 cases}
    \label{fig:box-cases}
    \bigskip

    \includegraphics[width=1.0\textwidth]{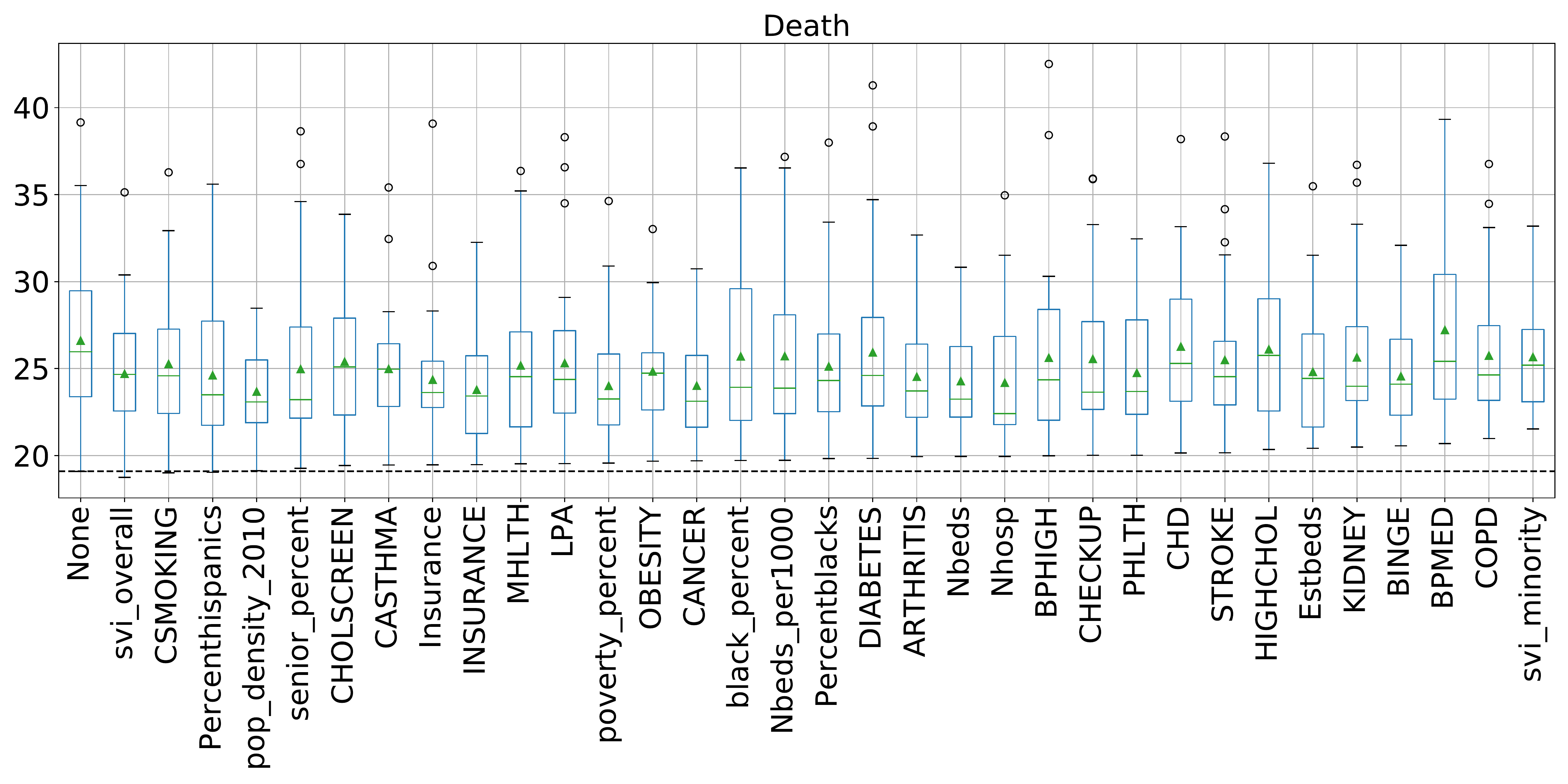}
    \vspace{-1cm}
    \caption{Influence of risk factors on the accuracy of the prediction of COVID-19 death}
    \label{fig:box-death}
\end{figure}

We have run the training on all 110 cities represented by the FIPS while not including any risk factors and running them with a single risk factor. We then summed up all absolute errors for the predicted data points over the same time period and repeated each experiment 10 times. We than agglomerated all results into a box-whisker diagram for death and for cases, respectively. We sorted the diagram in such a fashion that the risk factor that has the model with the lowest error comes first. We preceded the graph with the modes that did not use any risk factors to provide a comparison for minimizing the overall error. This calculation was done to identify risk factors that may have a better impact than others. The graphs are depicted in Figures \ref{fig:box-cases} and \ref{fig:box-death}. The order that we received for the best prediction models based on the risk factors for case and death are as follows:

\renewcommand{\_}{\textunderscore\hspace{0pt}}
\setlength{\emergencystretch}{3em}

\begin{description}

\item[Order of the cases sorted by minimum model error:] None, pop\_density\_2010, PHLTH, Insurance, Percentblacks, Percenthispanics, Nhosp, INSURANCE, CHOLSCREEN, DIABETES, STROKE, CHD, CHECKUP, Nbeds, svi\_overall, KIDNEY, CASTHMA, 
black\_percent, BPMED, LPA, CSMOKING, ARTHRITIS, poverty\_percent, Estbeds, MHLTH, senior\_percent, COPD, CANCER, Nbeds\_per1000, BPHIGH, HIGHCHOL, BINGE, OBESITY, svi\_minority

\item[Order of the death sorted by minimum model error:]
None, svi\_overall, CSMOKING, Percenthispanics, pop\_density\_2010, senior\_percent, CHOLSCREEN, CASTHMA, Insurance, INSURANCE, MHLTH, LPA, poverty\_percent, OBESITY, CANCER, black\_percent, Nbeds\_per1000, Percentblacks, DIABETES, ARTHRITIS, Nbeds, Nhosp, BPHIGH, CHECKUP, PHLTH, CHD, STROKE, HIGHCHOL, Estbeds, KIDNEY, BINGE, BPMED, COPD, svi\_minority
\end{description}

\begin{figure}[!p]
    \begin{minipage}{.45\textwidth}
    \centering
    \includegraphics[width=1.0\textwidth]{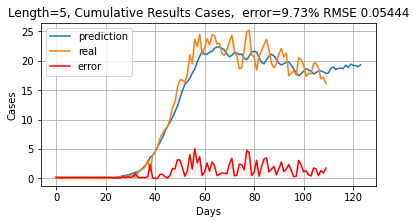}
    \vspace{-1cm}
    \caption{Model prediction error for cumulative cases when including one risk factor - pop\_density\_2010 in the features, in addition to the past number of cases and deaths. }
    \label{fig:error-case-popdensity}
    \end{minipage}
    \begin{minipage}{.45\textwidth}
    \ \
    \centering
    \includegraphics[width=1.0\textwidth]{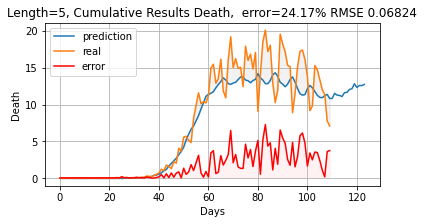}
    \vspace{-1cm}
    \caption{Model prediction error for cumulative deaths when including one risk factor - pop\_density\_2010 in the features, in addition to the past number of cases and deaths.}
    \label{fig:error-death-popdensity}
    \end{minipage}
\end{figure}

To show the values for the errors and their association with the risk factor, we include Table \ref{tab:top-1}, where cum\_error represents the cumulative error of the prediction. We sorted it by the cumulative error for cases. We see that in our experiment, almost all risk factors lead to an increased model prediction. As a result, we find that when including factors such as population density, physical health Insurance, population breakdown by ethnicity number of hospitals, and diabetes Leeds to better overall predictions for cases,
We also see that many of them are on average better,  

We are obtaining a different picture for the death, where we only identified two risk factors, namely the Social Vulnerability Index (SVI) and chain-smoking, that, when integrated into our model, leads to better prediction results. However, the accuracy of the overall prediction for death is far less precise than that of the cases. To showcase this, we have included Figures \ref{fig:error-case-popdensity} and \ref{fig:error-death-popdensity} that show the respective errors. This is following our input data distribution in which we find much higher fluctuations relative to the overall value. A mitigation to this issue would be using a seven day average over the analyzed period. However, we have not done this on purpose for this analysis as we wanted to identify how an LSTM behaves based on the daily fluctuations. This was done to avoid any needed prepossessing and to run and to identify the capability of the deep learning framework to adapt to such changes without any special activities. This is done to prepare our framework for an automated ingest of data on a daily basis for data fusion of new cases and death information. This way, we can train the model with new incoming data every day and make sure that the mode is kept state-of-the-art. Through this testing, we found that the hyperparameters such as epochs and dropout values work well for our experiments.

\begin{figure}[!p]
    \begin{minipage}{.45\textwidth}
        \centering
        \includegraphics[width=1.0\textwidth]{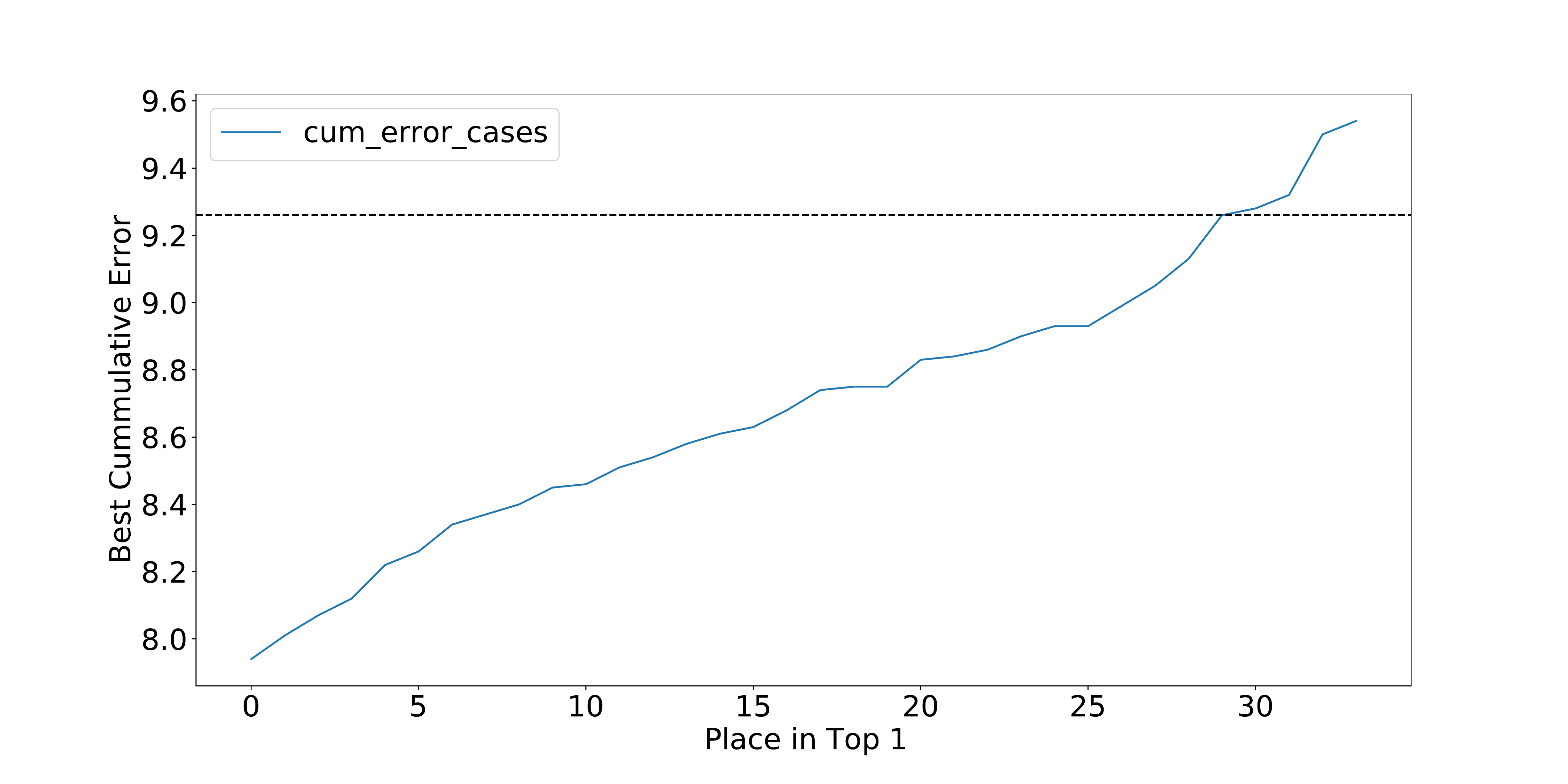}
        \vspace{-1cm}
        \caption{Place Top1 Cases }
        \label{fig:place-top1-cases}

    \end{minipage}
    \ \
    \begin{minipage}{0.45\textwidth}
        \centering
        \includegraphics[width=1.0\textwidth]{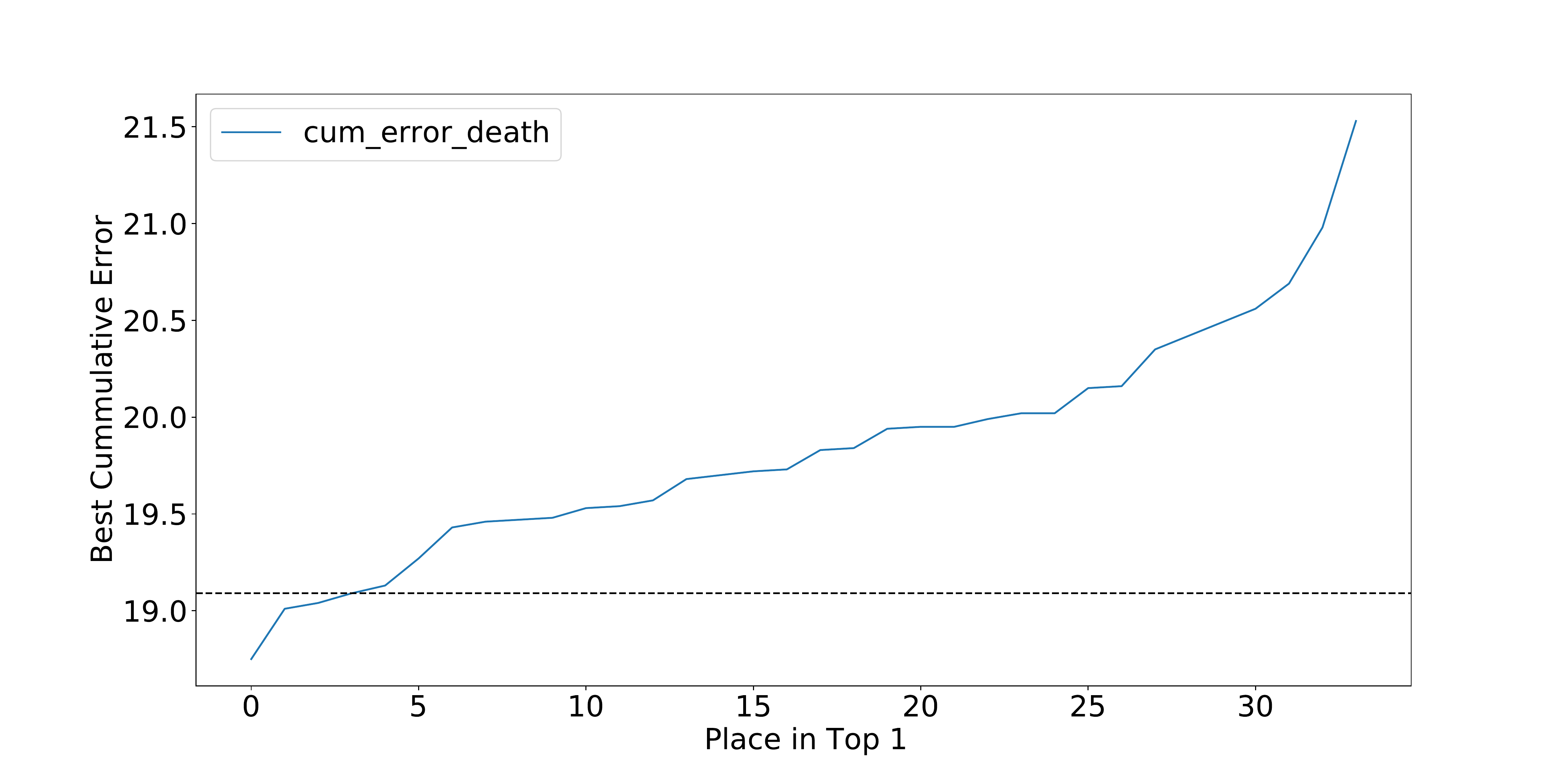}
        \vspace{-1cm}
        \caption{Place Top1 Death }
        \label{fig:place-top1-death}
    \end{minipage}

    \begin{minipage}{.45\textwidth}
        \centering
        \includegraphics[width=1.0\textwidth]{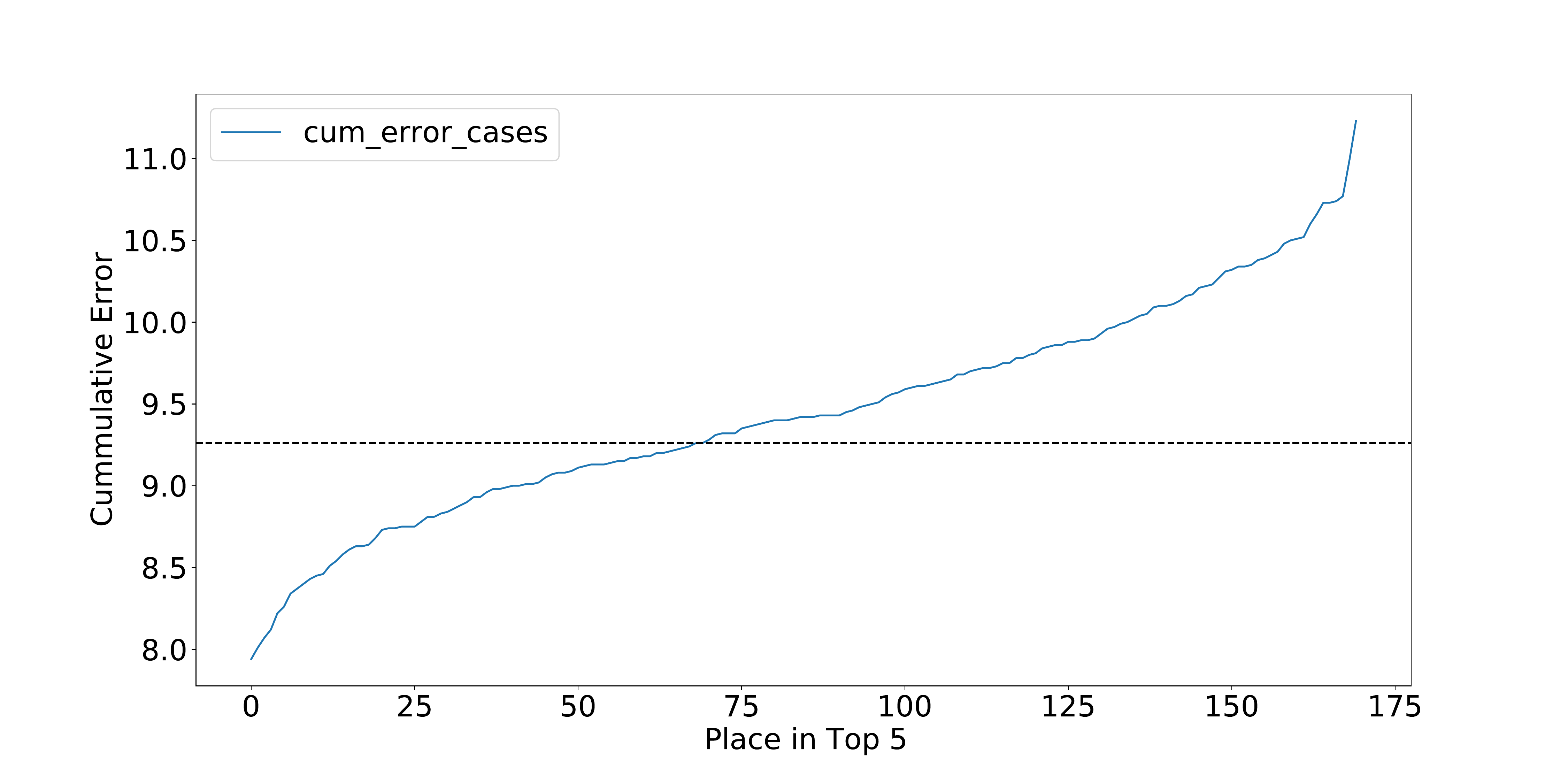}
        \vspace{-1cm}
        \caption{Place Top5 Cases }
        \label{fig:place-top5-cases}
    \end{minipage}
    \ \
    \begin{minipage}{.45\textwidth}
        \centering
        \includegraphics[width=1.0\textwidth]{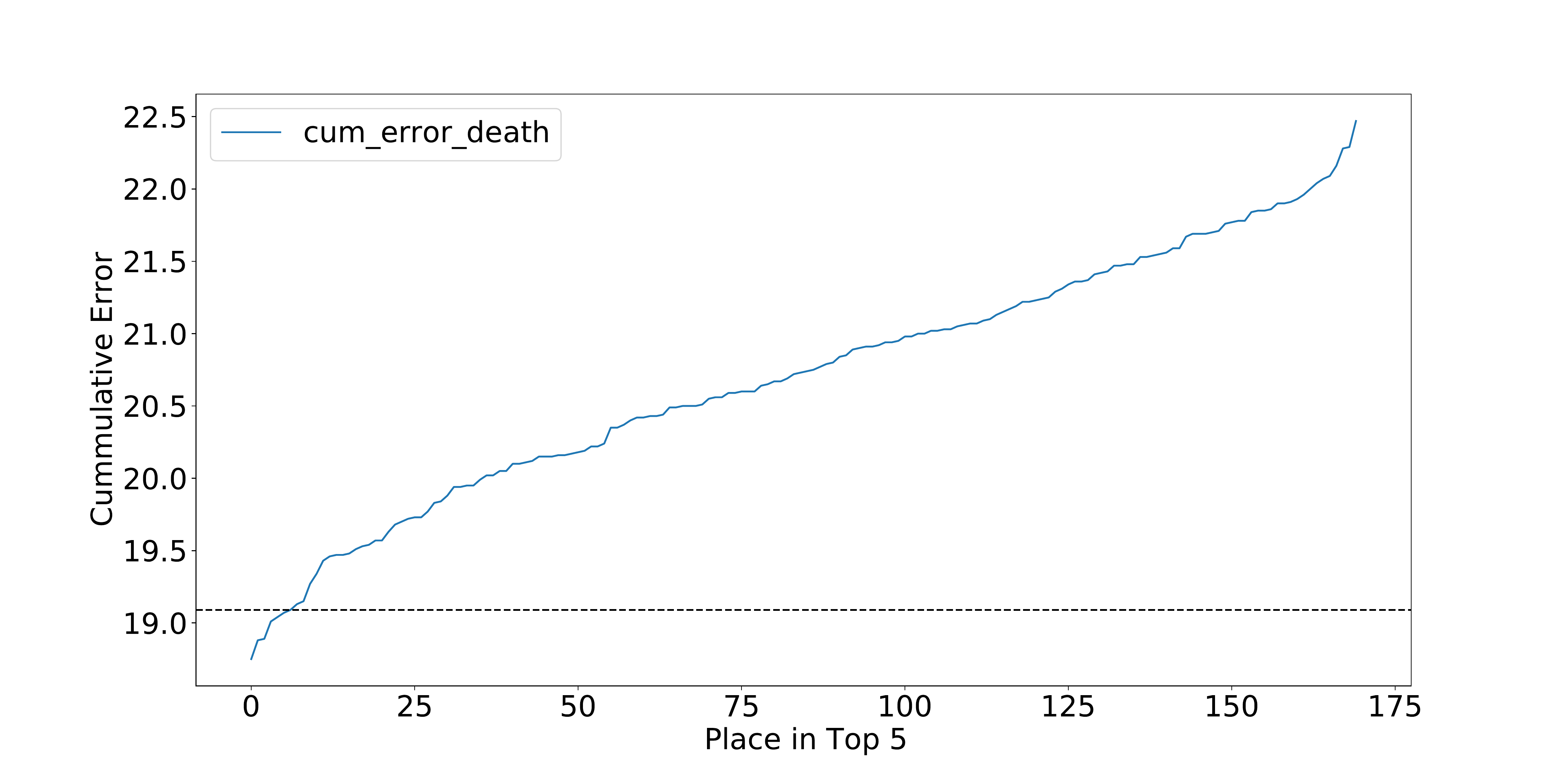}
        \vspace{-1cm}
        \caption{Place Top5 Death }
        \label{fig:place-top5-death}
    \end{minipage}

    \begin{minipage}{.45\textwidth}
        
        \centering
        \includegraphics[width=1.0\textwidth]{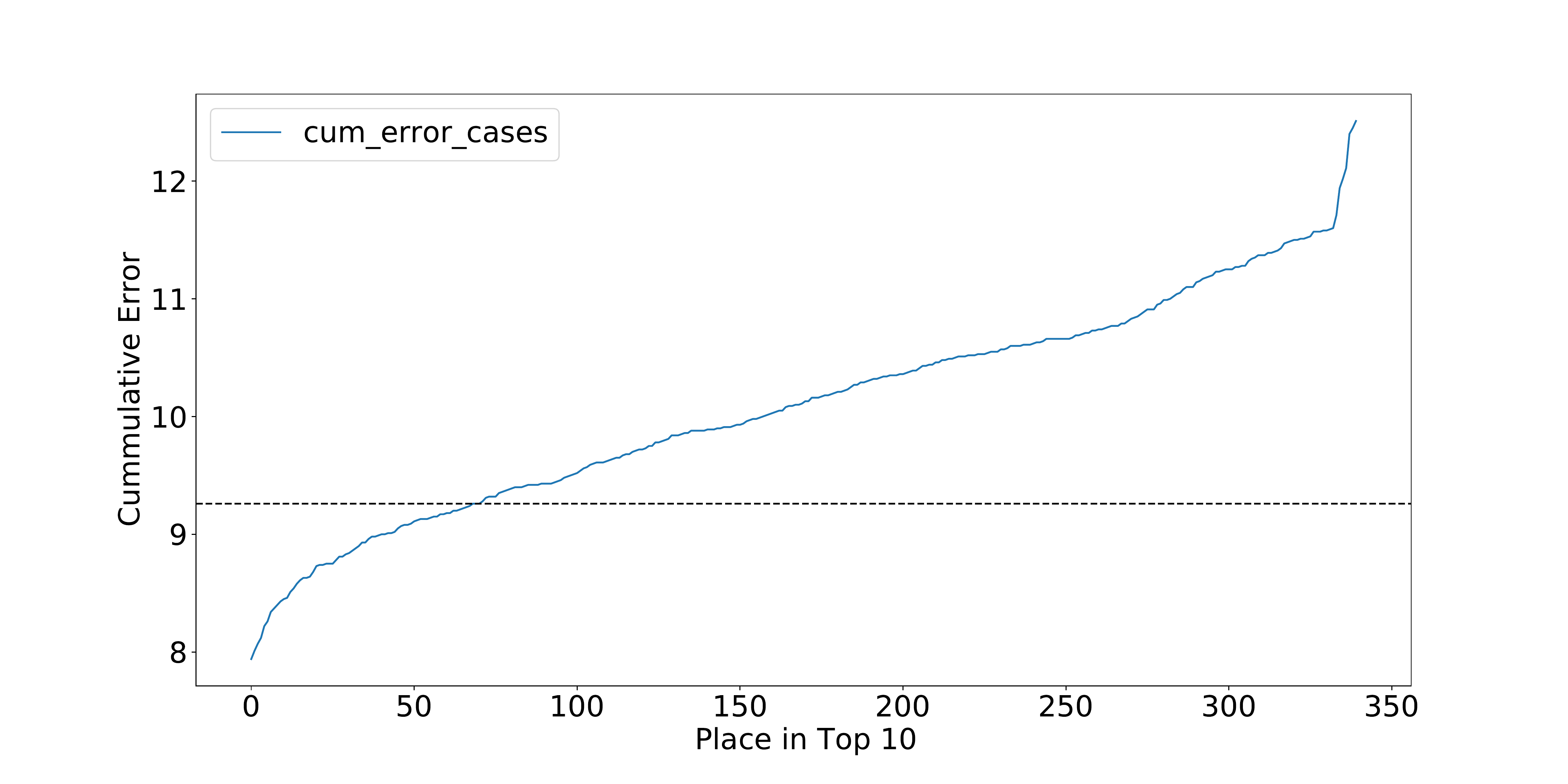}
        \vspace{-1cm}
        \caption{Place Top10 Cases }
        \label{fig:place-top10-cases}
    \end{minipage}
    \ \
    \begin{minipage}{.45\textwidth}
        
        \centering
        \includegraphics[width=1.0\textwidth]{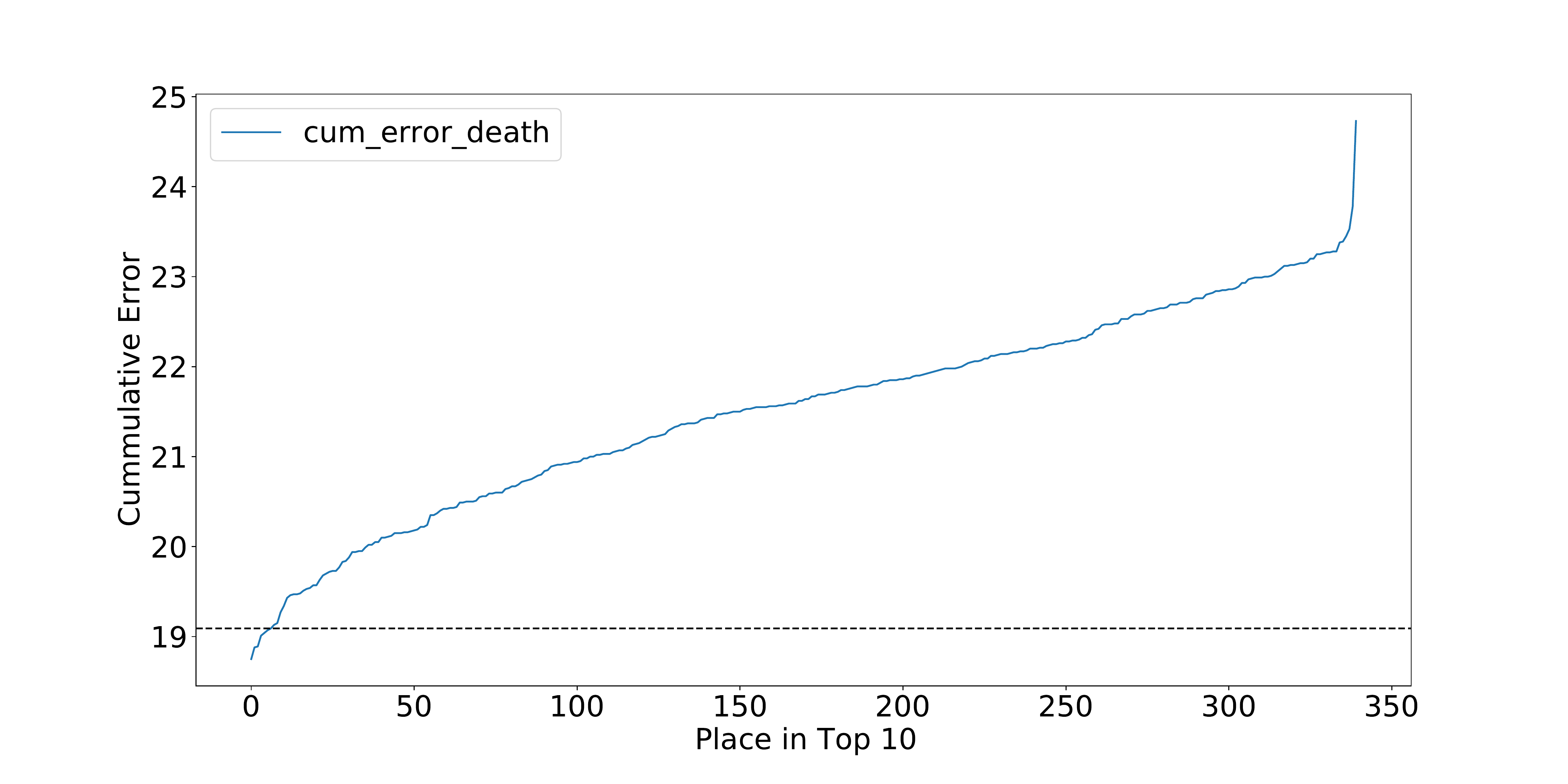}
        \vspace{-1cm}
        \caption{Place Top10 Death }
        \label{fig:place-top10-death}
    \end{minipage}
\end{figure}

We conducted one additional analysis to plot the best risk factor prediction model errors, the five best, and the ten best to identify if we have a
difficult time getting stuck in bad predictions. Thus we show Figures~\ref{fig:place-top1-cases} to  \ref{fig:place-top10-death}. As we can see, we identified that a jump occurs when we sort the top 10 model predictions at around 330 models from a total of 3400 automatically generated models. This also showcases that we must run our AICov model generator multiple times to obtain reasonable parameters resulting in us having to run our model generator for each factor at least ten times. This is a variable parameter to our model generator and can be adjusted.   

\begin{table}[p]
\caption{Best models for each risk factor and their respective errors}
\label{tab:top-1}
\bigskip
\centering
\resizebox{0.7\textwidth}{!}{ 
\begin{tabular}{rrrrrlrl}
\toprule
 rmse\_cases &  rmse\_death &  cum\_error\_cases &  cum\_error\_death &  days in & risk &  place & \\
\midrule
 0.055443 & 0.067617 & 7.94 & 22.74 & 5 & pop\_density\_2010 & 0 \\
 0.054795 & 0.068529 & 8.01 & 20.55 & 3 & PHLTH & 1 \\
 0.055199 & 0.067641 & 8.07 & 20.43 & 4 & Insurance & 2 \\
 0.055028 & 0.067207 & 8.12 & 20.17 & 4 & Percentblacks & 3 \\
 0.056061 & 0.067489 & 8.22 & 19.04 & 5 & Percenthispanics & 4 \\
 0.054621 & 0.068255 & 8.26 & 20.10 & 3 & Nhosp & 5 \\
 0.054557 & 0.068572 & 8.34 & 24.61 & 3 & INSURANCE & 6 \\
 0.054712 & 0.069121 & 8.37 & 21.70 & 3 & CHOLSCREEN & 7 \\
 0.054586 & 0.067636 & 8.40 & 20.50 & 5 & DIABETES & 8 \\
 0.054160 & 0.066906 & 8.45 & 20.16 & 4 & STROKE & 9 \\
 0.054027 & 0.067642 & 8.46 & 21.93 & 5 & CHD & 10 \\
 0.054927 & 0.067652 & 8.51 & 20.42 & 5 & CHECKUP & 11 \\
 0.054146 & 0.067536 & 8.54 & 21.15 & 5 & Nbeds & 12 \\
 0.054795 & 0.067676 & 8.58 & 19.94 & 5 & svi\_overall & 13 \\
 0.054600 & 0.067565 & 8.61 & 21.86 & 5 & KIDNEY & 14 \\
 0.055089 & 0.068248 & 8.63 & 22.58 & 5 & CASTHMA & 15 \\
 0.054547 & 0.068266 & 8.68 & 20.49 & 3 & black\_percent & 16 \\
 0.054685 & 0.068920 & 8.74 & 21.69 & 3 & BPMED & 17 \\
 0.054323 & 0.067149 & 8.75 & 19.54 & 5 & LPA & 18 \\
 0.053716 & 0.066898 & 8.75 & 19.01 & 5 & CSMOKING & 19 \\
 0.054053 & 0.068716 & 8.83 & 20.94 & 3 & ARTHRITIS & 20 \\
 0.055029 & 0.068672 & 8.84 & 22.02 & 3 & poverty\_percent & 21 \\
 0.054545 & 0.067267 & 8.86 & 21.22 & 4 & Estbeds & 22 \\
 0.054864 & 0.068742 & 8.90 & 19.53 & 3 & MHLTH & 23 \\
 0.055079 & 0.069024 & 8.93 & 21.23 & 3 & senior\_percent & 24 \\
 0.055589 & 0.068250 & 8.93 & 23.26 & 4 & COPD & 25 \\
 0.054500 & 0.068363 & 8.99 & 19.70 & 3 & CANCER & 26 \\
 0.054397 & 0.068491 & 9.05 & 21.89 & 3 & Nbeds\_per1000 & 27 \\
 0.054609 & 0.067346 & 9.13 & 21.29 & 4 & BPHIGH & 28 \\
 0.054868 & 0.069432 & 9.26 & 25.88 & 3 & \underline{None} & 29 \\
 0.056429 & 0.069345 & 9.28 & 20.98 & 3 & HIGHCHOL & 30 \\
 0.054317 & 0.067638 & 9.32 & 21.00 & 5 & BINGE & 31 \\
 0.054469 & 0.068540 & 9.50 & 20.65 & 3 & OBESITY & 32 \\
 0.055094 & 0.067443 & 9.54 & 21.53 & 4 & svi\_minority & 33 \\
\bottomrule
\end{tabular}
}
\end{table}

\clearpage

\section{Setting up the Deep Learning Framework}

We often think of the laws of physics described by operators that
evolve the system given sufficient initial conditions and in this
language, we have shown how to represent Newton’s law operator by a
recurrent network \cite{Kadupitiya2020-zq}. We expect that the time
dependence of many complex systems: Covid pandemics, Southern
California earthquakes, traffic flow, security events can be described
by deep learning operators that both capture the dynamics and allow
predictions. In this paper, we adopt this approach to describe the
time dependence of COVID-19 infection and fatality counts in different
regions. We start with a basic LSTM operator shown in table 1 with a
similar architecture to \cite{Kadupitiya2020-zq} with two layers of
LSTM and initial and final fully connected layers. We intend to
process the two time-series (infections, fatalities) associated with
each city plus the covariates (fixed in time but dependent on region)
to build a model for the time evolution of the data. There is a
nontrivial architecture design choice as to the method of handling the
time-independent covariates. We could combine a recurrent network for
the time series with some fixed network for the covariates. However,
that does not fit with our concept of deriving an evolution operator
for COVID-19 data that naturally depends on the covariates for each
city. Thus instead we feed the covariates into the recurrent network
as additional features leading to 35 of them if we use 33 covariates.
We did look at two related ways of doing this

\begin{description}

\item [a)]    Have covariates fixed in time

\item[b)] Multiply the covariates by the times series (we used the
  average of infections and fatalities)

\end{description}

We expected the second choice to be the most natural as it gives all
35 input streams similar time dependence. However, in our initial
tests, we found these choices gave very similar results and so we
fixed on the simpler choice a) in this paper. In the runs discussed
here, we either used 0, 33, or 1 covariate. Choosing 33 gives the best
description (reducing loss by about 1\%) while using each covariate as
a solitary extra feature allows us a rather sensitive study of the
importance of each individual covariate.

The final architecture choice was the quantities to predict (and train
on). We choose this to be both the infection/fatality values at the
time value following the sequence combined with the same quantities
for the following 14 days. This gives us 30 different output features
for each sequence. We renormalized so the two basic features
(following day) contributed 50\% of the loss function. We supported
this in TensorFlow with a custom mean square error ($MSE$) loss
function that dropped training values that were not available for time
sequences near the end of the examined period. Although one can’t
train on future values, one can predict them as seen in our sample
outputs. We extensively studied both 2 and 30 output features and they
give similar descriptions of the observed data with the 30 feature
choice enabling 14-day predictions.

We performed an extensive hyper-parameter search and the values
presented in Table 1 are good values that give robust results without
overfitting. We looked at four different activation functions in the
LSTM and fully connected layers. Selu, Relu and Tanh activations gave
very similar results while Sigmoid activation was much worse. We also
looked at the number of hidden units (separately for LSTM and fully
connected layers), the number of LSTM layers and the dropout values.
Only the latter product significant change and for example removing
dropouts would decrease the loss by about 16\% but we kept to provide
a robust environment. We chose randomly 20\% of the data for
validation and testing and this had similar loss values to the
training data (the other 80\%) with our chosen dropout.

The final technical issue concerns the data normalization where each
feature is transformed by a feature dependent but city independent way
to lie between 0 and 1. Further all extensive (proportional to size)
covariates were first divided by the city population except for
population itself where we took the logarithm; intensive covariates
were just rescaled to lie between 0 and 1. The basic COVID-19 daily
data is extensive but compared two approaches with and without
renormalization but in case we took the square root of the data value.
This is natural if there are counting (square root(N) for N
observations) errors as the deep learning loss function is
$(predicted-observed)^{2}$ without the error traditional in
$\chi^{2}$, namely $((predicted-observed)/error)^{2}$. This approach
with standard deep learning $MSE$ appears sound without dividing by
the population. That approach will emphasize the error contribution of
smaller cities but is still interesting to look at as there are
clearly many systematic (not counting) errors.

\section{Direct Deductions from Deep Learning Data Representation}

he deep learning model contains much information of which we have only
done a preliminary analysis as discussed already in Section
\ref{sec:lstm} and Figures \ref{fig:magic-1} and
\ref{fig:error-case-popdensity}. In addition, we have looked at the
sensitivity of the results as measured by the predicted daily rates
and the value of the loss function. In Table \ref{tab:corr-matrix}, we
present one way of looking at it. We calculate the correlation
coefficients of many of the covariates from an accurate LSTM model fit
to the data. We divide the time series into three regions

\begin{itemize}
\item \textbf{Past} or all days up to 14 days before the final observed date (May 25, 2020)
\item \textbf{Now} or the 14 days from May 12 to May 25, 2020
\item \textbf{Future} or the 14 days after May 25, 2020 predicted by the model
\end{itemize}
 
These results are different between population normed or direct
un-normalized data and here we keep the square-root used in the model
but divide model values by the square root of the population when
calculating correlation. Significant positive correlations are seen
for CASTHMA, HIGHCHOL, DIABETES, OBESITY, CSMOKING, CHD, and CHECKUP
while negative correlations for PVI and norm\_pop. We also looked at
the impact of these covariates on the evolution function. As this
operates on data at previous time values, this is sensitive to
variations across cities and times. The evolution operator is
intuitively the time derivative of the absolute values of daily rates
seen in the correlation in Table 3. Nhosp and Estbeds have the most
significance in impacting (as a reduction) the values of both the
infection and fatality rates from the evolution operator. The fatality
rates are also significantly impacted by Insurance and norm\_pop
(reduction) and senior\_percent and OBESITY (increase) from the
evolution operator. norm\_pop impacts infection values as an increase.
We can also identify which covariates have the most significant impact
on the loss function. Here Nhosp, CHD, KIDNEY, BINGE, PHLTH give 5\%
or more reduction in loss function while norm\_pop, svi\_overall,
CASTHMA, DIABETES, BPHIGH have small (1\%) effects.

\begin{table}[!h]
\caption{Correlation Matrix}
\bigskip
\label{tab:corr-matrix}
\resizebox{1.0\textwidth}{!}{
\begin{tabular}{lrrrrrr}
\toprule
Risk Factor		& Past Case & Now Case & Future Case & Past Death & Now Death & Future Death \\
\midrule
PVI                     & -0.425      & -0.292      & -0.124      & -0.434      & -0.373      & -0.239    \\
Insurance               & 0.018       & 0.017       & 0.012       & 0.018       & -0.002      & 0.001     \\
Nbeds                   & 0.136       & 0.131       & 0.217       & 0.089       & 0.080        & 0.109     \\
Nbeds\_per1000           & 0.138       & 0.121       & 0.209       & 0.092       & 0.069       & 0.088     \\
Nhosp                   & 0.074       & 0.119       & 0.165       & -0.009      & 0.028       & 0.021     \\
Estbeds                 & 0.136       & 0.131       & 0.217       & 0.089       & 0.08        & 0.109     \\
senior\_percent          & 0.104       & 0.028       & 0.033       & 0.152       & 0.164       & 0.156     \\
black\_percent           & 0.209       & 0.099       & 0.208       & 0.216       & 0.085       & 0.116     \\
Percenthispanics        & -0.072      & -0.032      & -0.064      & -0.091      & -0.068      & -0.046    \\
pop\_density\_2010        & -0.302      & -0.21       & -0.175      & -0.272      & -0.231      & -0.216    \\
poverty\_percent         & 0.128       & 0.059       & 0.2         & 0.148       & 0.031       & 0.08      \\
svi\_minority            & 0.029       & 0.069       & 0.081       & 0.01        & 0.006       & 0.087     \\
svi\_overall             & 0.14        & 0.158       & 0.238       & 0.125       & 0.113       & 0.179     \\
CASTHMA                 & 0.238       & 0.271       & 0.307       & 0.205       & 0.288       & 0.309     \\
HIGHCHOL                & 0.202       & 0.185       & 0.131       & 0.192       & 0.243       & 0.224     \\
DIABETES                & 0.200         & 0.200         & 0.178       & 0.206       & 0.232       & 0.23      \\
OBESITY                 & 0.199       & 0.205       & 0.194       & 0.191       & 0.229       & 0.212     \\
CANCER                  & -0.145      & -0.102      & -0.129      & -0.138      & -0.058      & -0.059    \\
STROKE                  & 0.123       & 0.125       & 0.117       & 0.167       & 0.143       & 0.135     \\
MHLTH                   & 0.11        & 0.161       & 0.186       & 0.076       & 0.175       & 0.222     \\
CSMOKING                & 0.189       & 0.258       & 0.249       & 0.16        & 0.289       & 0.323     \\
CHOLSCREEN              & 0.164       & 0.001       & -0.026      & 0.187       & 0.07        & 0.038     \\
INSURANCE               & 0.018       & 0.017       & 0.012       & 0.018       & -0.002      & 0.001     \\
CHD                     & 0.225       & 0.284       & 0.233       & 0.178       & 0.327       & 0.337     \\
CHECKUP                 & 0.206       & 0.221       & 0.28        & 0.201       & 0.272       & 0.343     \\
KIDNEY                  & 0.045       & 0.028       & 0.032       & 0.022       & 0.071       & 0.053     \\
BINGE                   & 0.031       & 0.044       & 0.009       & 0.083       & 0.03        & 0.022     \\
LPA                     & 0.191       & 0.187       & 0.197       & 0.147       & 0.22        & 0.214     \\
ARTHRITIS               & 0.046       & 0.013       & -0.021      & 0.061       & 0.089       & 0.079     \\
BPMED                   & 0.060        & 0.080        & 0.124       & 0.056       & 0.073       & 0.09      \\
PHLTH                   & 0.064       & 0.132       & 0.104       & 0.022       & 0.129       & 0.114     \\
BPHIGH                  & 0.056       & 0.051       & 0.048       & 0.058       & 0.084       & 0.068     \\
COPD                    & 0.163       & 0.163       & 0.153       & 0.175       & 0.22        & 0.24      \\
norm\_pop                & -0.305      & -0.258      & -0.218      & -0.193      & -0.218      & -0.099  \\
\bottomrule
\end{tabular}
}
\end{table}
 
\section{Conclusion and Future Work}

We present AICov as an integrative deep learning framework for COVID-19 forecasting with population covariates. One of the important features of the Architecture for AICov is that it is by design targeting Cloud and HPC resources to conduct parameter sweeps to leverage sophisticated deep learning toolkits. The architecture allows the integration of various data sources that can update the data from its sources on demand and update its model predictions based on new data being injected. Parameters can easily be adjusted via Jupyter notebooks that intern call the computational backends on HPC and cloud resources.

In addition to this architecture, we used data collected from multiple public sources and agencies, and integrated the same across spatially contiguous units such as cities or metropolitan areas. In our pilot study, we have focused on 110 selected cities of the US, but the framework is general and can be used for analysis of other worldwide data.

The analysis we have conducted can be understood as a pilot to identify if it is feasible to be used. From this pilot, we deduce it actually can lead to an improved prediction once we integrate risk factors in addition to the time-dependent data such as cases and death resulting from COVID-19. While the improvements we have observed are modes, they do present a way of improving the forecast models. Inclusion of further putative factors from the ongoing worldwide studies on Covid-19 will only strengthen the future applications of our integrative framework.

We have shown that deep learning can return very good results while using smooth data as we used in our empirical fits. For real data as presented to us for the daily changes, it still produced good results. Also, we have used in our data only cases and death, but we intend to expand this by using data about recovery.

We have experimented with different hyperparameters and included in this study a selection of hyperparameters that have worked well for this data set. 

As we have set up the first version of our AICov software, we will improve it to include high-performance computers instead of just cloud resources to leverage the universities' compute infrastructure. Furthermore, we are currently conducting experiments with multiple risk factors at the same time to identify the most significant combinations of them.
As we have shown that the bivariate analysis has a locality dependency, we will explore categorizing locations with similar demographics and conduct our experiments on demo-graphical similar regions. We will improve the implementation with the goal of open-sourcing the software. 

\vskip 14pt
\noindent {\bfseries{\fontsize{14pt}{1em}\selectfont Acknowledgements}}

This work is partially supported by the National Science Foundation (NSF) through awards CIF21 DIBBS 1443054, nanoBIO 1720625, Cybertraining 1829704, CINES 1835598 and Global Pervasive Computational Epidemiology 1918626.  We thank Jcs Kadupitiya for several useful discussions on deep learning frameworks.
\par

\clearpage

\section {Abbreviations}

We are using the following abbreviations in the paper and the data with data from \cite{cdc-brfss}, 
\cite{cdc-svi} and \cite{www-500Citie16} of population $>=18$ years of age that have been data fused by University of Pittsburgh.

\begin{table}[!h]
\caption{Risk factor abbreviations}
\label{tab:risk-factors}
\bigskip
\resizebox{0.90\textwidth}{!}{
\centering
\begin{tabular}{ll}
\toprule
Risk Factor & Description \\
\midrule
 PHLTH        & Percent of Physical Health\\
 ARTHRITIS 	  & Percent of arthritis \\
 BINGE 		  & Percent of binge drinking \\
 BPHIGH 	  & Percent of high blood pressure\\
 BPMED 		  & Percent of taking blood pressure medication\\
 CANCER 	  & Percent of cancer patients\\
 CASTHMA 	  & Percent of current asthma\\
 CHD 		  & Coronary heart disease\\
 CHECKUP 	  & Percent of health checkup\\
 CHOLSCREEN   & Percent of cholesterol screening\\
 COPD 		  & Percent of Chronic obstructive pulmonary disease\\
 CSMOKING 	  & Current smoking in percent\\
 DIABETES 	  & Percent of diabetes\\
 Estbeds 	  & Number of estimated beds in hospitals\\
 HIGHCHOL 	  & Percent of high cholesterol\\
 Insurance 	  & Percent of Insurance\\
 KIDNEY 	  & Percent of kidney disease\\
 LPA 		  & Percent of No leisure-time physical activity \\
 MHLTH 		  & Percent of not good mental health\\
 Nbeds 		  & Number of beds in hospitals \\
 Nbeds\_per1000   & Number of beds per 1000 people\\
 Nhosp 		  & Number of hospitals\\
 OBESITY 	  & Percent of population that is obese\\
 PHLTH		  & Percent of Physical Health\\
 PVI 		  & Pandemic vulnerability index \\
 Percentblacks 	  & Percent of Blacks in the population\\
 Percenthispanics & Percent of Hispanics in the population\\
 STROKE 	  & Percent of stroke\\
 \underline{None} & no risk factor used\\
 black\_percent	  & Percent of blacks in the population\\
 poverty\_percent & Percent of poverty in the population\\
 senior\_percent  & Percent of seniors in the population\\
 svi\_minority 	  & Percent of social vulnerability index in the minority population\\
 svi\_overall 	  & Percent of social vulnerability index in the overall population\\
\bottomrule
\end{tabular}
}
\end{table}

\clearpage

\bibliographystyle{IEEEtran}

\bibliography{ms}

\end{document}